\documentclass[twocolumn,11pt, a4paper]{article}
\usepackage[fleqn]{amsmath}
\usepackage{epsfig}
\usepackage{times}
\usepackage{cite}
\usepackage{lineno,amsmath}
\usepackage[colorlinks=true, urlcolor=blue]{hyperref}
\usepackage[psamsfonts]{amssymb}
\usepackage{color}
\usepackage{subfigure}
\usepackage{subcaption}
\usepackage{booktabs}%
\usepackage{threeparttable}%
\usepackage{multirow}
\usepackage{soul, color, xcolor}

\usepackage{colortbl}
\AddToHook{env/figure/begin}{
  \ifLineNumbers\captionsetup{font+=linenumbers}\fi
}

\AddToHook{env/table/begin}{
  \ifLineNumbers\captionsetup{font+=linenumbers}\fi
}
\begin{document}
\global\def\refname{{\normalsize \it References:}}
\baselineskip 12.5pt
%
%
%
\title{\LARGE \bf Edge AI-Enabled Chicken Health Detection Based on Enhanced FCOS-Lite and Knowledge Distillation}

\date{}

\author{\hspace*{-40pt}
\begin{minipage}[t]{2.7in} \normalsize \baselineskip 12.5pt
\centerline{Qiang Tong$^{a}$}
\centerline{\it Research $\&$ Development}
\centerline{\it Center China}
\centerline{SONY Group Corporation}
\centerline{Beijing, CHINA}
\centerline{$^a$\bf Corresponding Author}
\end{minipage} \kern -0.5in
\begin{minipage}[t]{2.7in} \normalsize \baselineskip 12.5pt
\centerline{Jinrui Wang}
\centerline{\it School of Computer Science}
\centerline{Beijing University of }
\centerline{Posts and Telecommunications}
\centerline{Beijing, CHINA}
\end{minipage} \kern -0.5in
\begin{minipage}[t]{2.7in} \normalsize \baselineskip 12.5pt
\centerline{Wenshuang Yang}
\centerline{\it School of Electrical and}
\centerline{\it Electronic Engineering}
\centerline{Nanyang Technological University}
\centerline{SINGAPORE}
\end{minipage}
\\ \\ \hspace*{-40pt}
\begin{minipage}[t]{2.7in} \normalsize \baselineskip 12.5pt
\centerline{Songtao Wu}
\centerline{\it Research $\&$ Development}
\centerline{\it Center China}
\centerline{SONY Group Corporation}
\centerline{Beijing, CHINA}
\end{minipage} \kern -0.5in
\begin{minipage}[t]{2.7in} \normalsize \baselineskip 12.5pt
\centerline{Wenqi Zhang}
\centerline{\it Research $\&$ Development}
\centerline{\it Center China}
\centerline{SONY Group Corporation}
\centerline{Beijing, CHINA}
\end{minipage} \kern -0.5in
\begin{minipage}[t]{2.7in} \normalsize \baselineskip 12.5pt
\centerline{Chen Sun}
\centerline{\it Research $\&$ Development}
\centerline{\it Center China}
\centerline{SONY Group Corporation}
\centerline{Beijing, CHINA}
\end{minipage} 
\\\\\hspace*{-40pt}
\begin{minipage}[t]{2.7in} \normalsize \baselineskip 12.5pt
\centerline{Kuanhong Xu}
\centerline{\it Research $\&$ Development}
\centerline{\it Center China}
\centerline{SONY Group Corporation}
\centerline{Beijing, CHINA}
\end{minipage}
%
%
\\ \\ \hspace*{-10pt}
\begin{minipage}[b]{6.9in} \normalsize
\baselineskip 12.5pt {\it Abstract:}
Edge-AI based AIoT technology in modern poultry management has shown significant advantages for real-world scenarios, optimizing farming operations while reducing resource requirements. To address the challenge of developing a highly accurate edge-AI enabled detector that can be deployed within memory-constrained environments, such as a highly resource-constrained edge-AI enabled CMOS sensor, this study innovatively develops an improved FCOS-Lite detector as a real-time, compact edge-AI enabled detector designed to identify chickens and assess their health status using an edge-AI enabled CMOS sensor. The proposed FCOS-Lite detector leverages MobileNet as the backbone to achieve a compact model size. To mitigate the issue of reduced accuracy in compact edge-AI detectors without incurring additional inference costs, we propose a gradient weighting loss function for classification and introduce a CIOU loss function for localization. Additionally, a knowledge distillation scheme is employed to transfer critical information from a larger teacher detector to the FCOS-Lite detector, enhancing performance while preserving a compact model size. Experimental results demonstrate the proposed detector achieves a mean average precision (mAP) of 95.1 $\%$ and an F1-score of 94.2 $\%$, outperforming other state-of-the-art detectors. The detector operates efficiently at over 20 FPS on the edge-AI enabled CMOS sensor, facilitated by int8 quantization. These results confirm that the proposed innovative approach leveraging edge-AI technology achieves high performance and efficiency in a memory-constrained environment, meeting the practical demands of automated poultry health monitoring, offering low power consumption and minimal bandwidth costs.
\\ [4mm] {\it Key--Words:}
AIoT, Edge-AI enabled CMOS sensor, Chicken healthy status detection, FCOS-Lite, Knowledge distillation
\end{minipage}
\footnote{The updated version of this manuscript has been published in \textit{Computers and Electronics in Agriculture} (2024), Volume 226, Article 109432. DOI: \url{https://doi.org/10.1016/j.compag.2024.109432}.}
\vspace{-10pt}}
\maketitle

\thispagestyle{empty} \pagestyle{empty}
%
%
\section{Introduction}
\label{sec1} \vspace{-4pt}

Traditional approaches to chicken poultry welfare management are plagued by high labor costs and inefficient resource management such as power consumption \cite{intro1}. Monitoring poultry health is especially challenging, as continuous, efficient, and precise inspection by human workers is unfeasible for the thousands to millions of birds typically found on a poultry ranch. Recently, AIoT (AI and IoT) technologies have emerged as promising solutions to these challenges \cite{intro3, intro2}. AIoT can facilitate efficient resource control, significantly reduce the workload of human workers, and enhance overall farming efficiency by automating \cite{intro4} and optimizing poultry health monitoring \cite{intro19, intro5, intro6}

Edge computing, a key component of AIoT, has revolutionized practical applications by integrating network, computing, storage, and application capabilities into compact devices. Unlike high-performance computing equipment or cloud servers, edge computing devices are designed for low power consumption and minimal bandwidth usage while delivering services closest to the data source. This makes edge devices ideal for real-world scenarios that demand portability and efficiency. In this study, we utilize an edge AI-enabled CMOS sensor—IMX500 \cite{iot2} as our edge device. Unlike other GPU-based edge devices used in poultry farming \cite{intro17}, the CMOS sensor offers end users a highly cost-effective and simplified deployment solution, thanks to its significantly lower power consumption and compact size. However, the limited memory space (8MB) of the CMOS sensor poses significant challenges in developing an edge-AI detector that remains compact yet performs well in practical AIoT applications. Therefore, the objective of this study is to develop a real-time, compact edge-AI model that delivers  strong performance while operating with minimal computing power on a highly resource-constrained yet cost-efficient edge-AI enabled CMOS sensor. This approach aims to automate chicken health status detection by leveraging edge-AI technology in a novel way, addressing practical challenges and meeting real-world demands.

In the past years, with the advancement of deep learning-based object detection technologies, significant progress has been made in identifying the status of poultry. For instance, Yi Shi et al. \cite{intro10} proposed a lightweight YOLOv3-based detection network tailored for chicken recognition and monitoring. Similarly, Zhuang et al. \cite{intro11} proposed an improved SSD (Single Shot MultiBox Detector) model to detect and classify healthy and sick chickens in real time, which is a good example of utilizing object detection networks for real-time detecting chickens and recognizing their healthy statuses. Liu et al. \cite{intro12} designed and implemented a compact removal system designed for detecting and removing deceased chickens within poultry houses, leveraging the YOLOv4 network. Moreover, the authors in \cite{intro13, intro14} expanded the system's application to chicken houses with caged chickens, utilizing the networks based on YOLOv3 and YOLOv5 respectively, to distinguish between healthy and sick chickens. Furthermore, the authors in \cite{intro20} proposed a defencing algorithm based on U-Net to mitigate the effects of cage fences, thereby enhancing the accuracy of chicken detection using YOLOv5. Additionally, authors in \cite{intro21, intro22} introduced chicken detection methods based on U-Net to address challenges in crowded scenes and to identify chicken behaviors, respectively. These researches show the capability of object detection techniques in poultry health monitoring. However, the methods mentioned above have not primarily focused on developing edge-AI enabled detectors, thereby restricting their applicability to large machines with GPUs such as PC, server, or Nvidia Jetson Xavier-type machines. This constraint severely impedes the utilization of lightweight edge devices with low power consumption and minimal bandwidth usage in practical AIoT scenarios. 
\begin{figure*}[htbp]
\centering
\includegraphics[width=\textwidth]{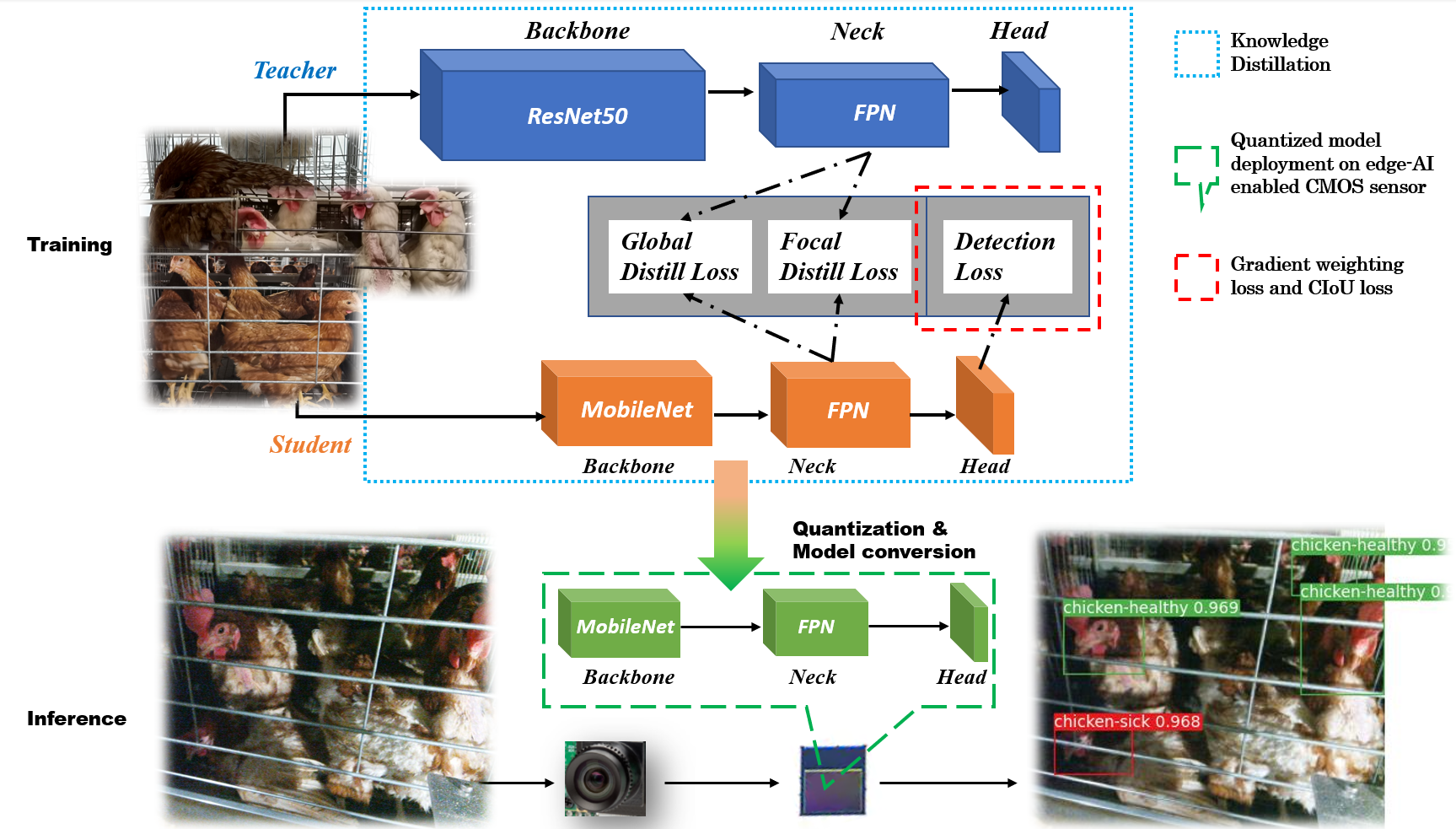}
\caption{Schematic of the edge-AI enabled detector. During the training phase, the compact FCOS-Lite detector, acting as the student model, is improved through knowledge distillation and tailored detection loss functions, then following compression for inference, the refined student model is deployed on the edge-AI enabled CMOS sensor.}
\label{fig1}
\end{figure*}

Several lightweight YOLO-based AI models, nano-level model and frameworks have been developed to address practical usage issues. For instance, in \cite{intro23}, the authors proposed an accurate method for chicken flock detection using the lightweight YOLOv7-tiny model. In \cite{intro16}, Knowledge Distillation (KD) techniques were employed to enhance the performance of the small YOLOv5s model for sheep face recognition, using the larger YOLOv5x model as the teacher detector. This approach effectively improves the performance of the compact model without increasing its size and inference costs. Additionally, in \cite{intro24}, a compact YOLO-Spot model was introduced for weed detection, leveraging edge computing devices. Moreover, in \cite{intro18}, authors presented the RTFD algorithm, based on PicoDet-S, for lightweight detection of fruits such as tomatoes and strawberries on edge CPU computing devices. Although the aforementioned methods offer lightweight solutions, they still struggle to achieve a good balance between accuracy and compact model size due to their reliance on anchor boxes or overly reduced model sizes. In contrast, FCOS \cite{dl1} stands out by delivering high accuracy, particularly in detecting objects of varying sizes and shapes, owing to its elimination of anchor boxes, simplified design, reduced need for extensive hyper-parameter tuning, and the use of decoupled detection heads. Additionally, the architecture of FCOS allows for adjustment of backbone to accommodate various model sizes, making the creation of an edge-AI version of FCOS a promising endeavor. Furthermore, our preliminary experiments and numerous existing studies have shown that FCOS with different ResNet backbones perform well in knowledge distillation, indicating that enhancing the performance of an edge-AI version of FCOS through knowledge distillation is also highly promising. While model pruning is effective in reducing model size for edge deployment, as demonstrated in \cite{intro15}, overly aggressive pruning of a large, accurate model to fit edge-AI constraints can significantly degrade accuracy if not carefully managed. Therefore, we select knowledge distillation as our technical approach to create an edge-AI enabled detector with good performance.

The key contributions in this study, regarding our proposed edge-AI enabled detector, are summarized as follows:
\begin{itemize}
\item We introduce a FCOS-Lite detector that utilizes MobileNet as the backbone and integrates modified neck and head components, resulting in a lightweight and compact model size suitable for edge-AI processing.
\item We propose a gradient weighting loss function and introduce CIOU loss function as the classification loss and localization loss respectively, aiming to enhance the accuracy of the proposed edge-AI enabled FCOS-Lite detector. Especially, the gradient weighting loss automatically assigns lower weights to easy samples and “outlier” samples, redirecting focus to other samples and thereby improving classification accuracy. 
\item We propose a knowledge distillation scheme to transfer valuable information from a large teacher model, such as the original FCOS detector with a ResNet backbone, to the proposed FCOS-Lite model. This approach effectively mitigates the accuracy reduction inherent in the compact FCOS-Lite model without additional inference costs. As a result, a favorable balance between high accuracy and a compact model size is achieved.  
\end{itemize}

The overview schematic of the proposed edge-AI enabled detector is shown in Fig.\ref{fig1}. During the training phase, our compact FCOS-Lite detector serves as the student model within the knowledge distillation scheme. The accuracy of the proposed detector is enhanced through the "knowledge transfer" from a larger teacher model and specifically designed detection loss functions. Then following additional model compression techniques, such as int8 quantization, the refined detector is deployable within the memory-constrained edge-AI enabled CMOS sensor for inference. By utilizing the edge-AI enabled CMOS sensor into a lightweight monitoring camera, our detection system guarantees low power consumption and minimal bandwidth costs, thereby ensuring cost-efficient practical applicability in AIoT scenarios.
 \vspace{-11pt}
\section{Materials and methods}\label{sec2}
 \vspace{-8pt}
 \begin{figure*}[htbp]
\centering
\includegraphics[width=\textwidth]{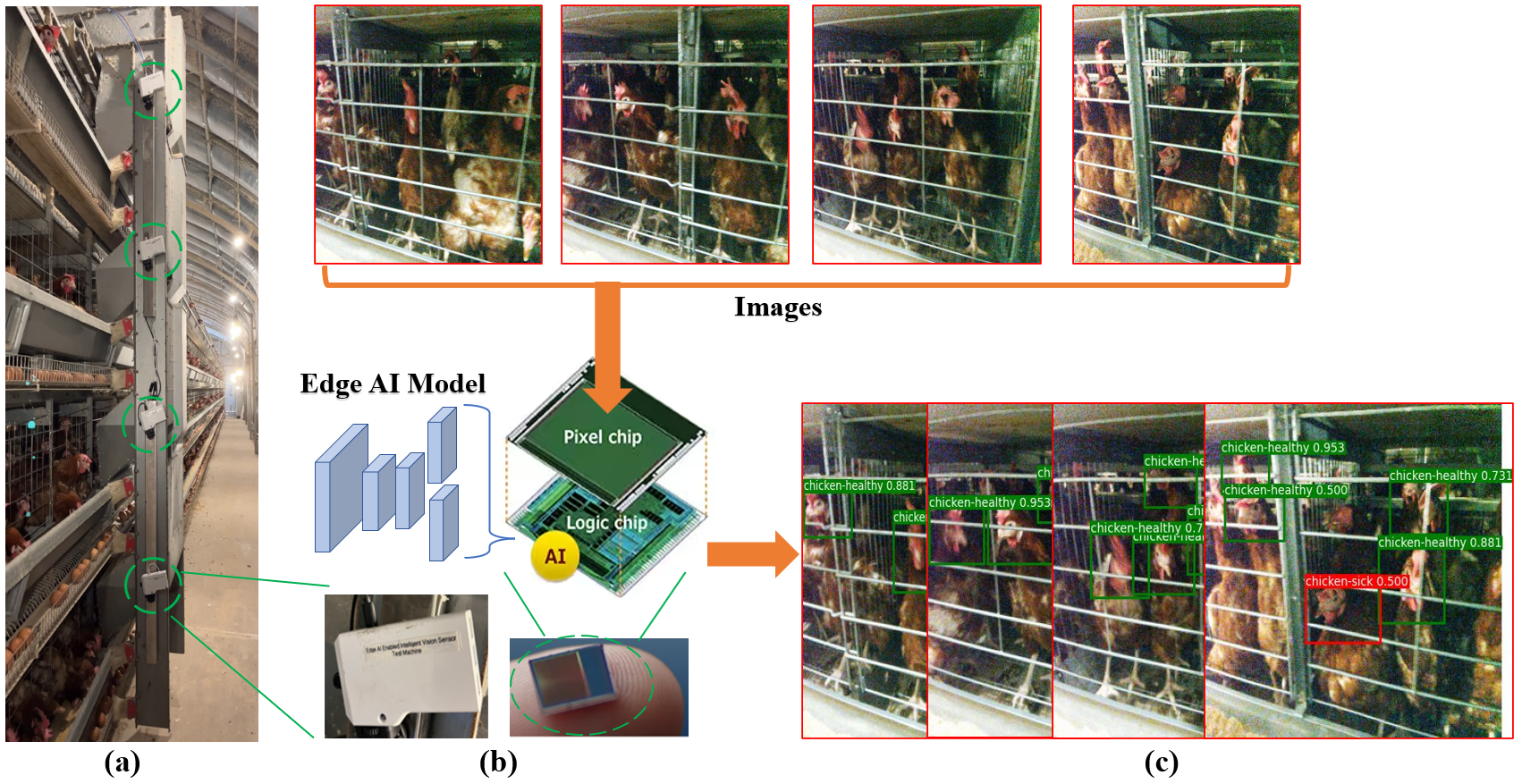}
\caption{Example of the whole system featuring light-weighted intelligent cameras and our proposed detector: (a) Overall system placement in a real-world AIoT scenario, (b) Intelligent camera (left) and its internal edge-AI enabled CMOS sensor (right), and (c) Example of a visual result outputted by the proposed detector.}
\label{fig2}
\end{figure*}
In this section, we first introduce the details of our experimental environment and the specific application scenario targeted by our proposed method in Sec.\ref{sec2.1}. Then we introduce the details of our proposed detector in Sec.\ref{sec2.2}.
\subsection{Materials}
\label{sec2.1}
\subsubsection{AIoT Scenario and edge device}
\label{sec2.1.1}
Fig.\ref{fig2} (a) shows an example of the experimental system, positioned on the automatic feeder located within the layer house. The automatic feeder autonomously moves along the length of the cage shield, dispensing feed to the enclosed chickens several times throughout the day. Four light-weighted intelligent cameras (inside green circles) are mounted on the automatic feeder, enabling autonomous and intelligent surveillance of the health statuses of chickens within the four-level cage arrays. An example of the external configuration of the intelligent camera along with its internal edge-AI enabled CMOS sensor are shown in the left and right sides in Fig.\ref{fig2} (b) respectively. In this study, we employ "IMX500" with an 8 MB memory capacity and a maximum computational power of 1 TOPS (Tera Operations Per Second) computational power for int8 data processing, as the edge-AI enabled CMOS sensor. This sensor incorporates a stacked structure (refer to the right side of (b)), integrating a regular image sensor, a robust DSP (Digital Signal Processor), and dedicated on-chip SRAM (Static Random-Access Memory) to facilitate accelerated edge-AI processing at impressive speeds. During the patrolling activities of the automatic feeder, the proposed edge-AI enabled detector deployed on the logic chip directly uses the frames captured by the pixel chip, as inputs, then automatically detects the locations of the chickens and identifies their respective health statuses. The outputs of the intelligent camera consist of metadata derived from edge-AI processing, such as the recognition of chicken health status (healthy or unhealthy) in tensor vector format. Alternatively, the outputs can also comprise images directly captured by the camera or can comprise visual results of ROI (Region Of Interest) on the captured images. Fig.\ref{fig2} (c) shows an example of visual output from the proposed detector deployed on the CMOS sensor. However, it's important to note that,  such visual output as shown in Fig.\ref{fig2} (c) for reporting visual recognition outcomes for all chickens, may not be essential in practical AIoT scenarios. Because of the intelligent camera's capability to execute edge-AI algorithms directly on the CMOS sensor, the outcomes of chicken health status monitoring, which are subsequently transmitted to higher-level computers such as the cloud servers, can be optimized to encompass metadata that solely includes byte-format messages of the recognition results for "unhealthy" chickens. As a result, the transmitted outputs from each intelligent camera are compact, consisting of just a few bytes. Therefore, during the system's daily patrol inspections, uploading outputs to the upper-level computer requires minimal cumulative bandwidth consumption and a low bit rate. Furthermore, within the intelligent camera shown in Fig.\ref{fig2} (b), in conjunction with lightweight and low power consumption micro boards such as "Raspberry Pi model 4B", "Raspberry Pi Zero 2W", etc., the camera's power ($\sim$5V, $\sim$1.5A) can be supplied through a USB connection, utilizing a mobile power supplement or a battery pack (2.5V $\sim$ 5V) as the energy source. As a conclusion, the edge-AI enabled CMOS sensor based intelligent camera can effectively minimize bandwidth costs and allows the entire patrolling system to leverage benefits of low power consumption and reduced bandwidth costs, making it well-suited for real-world AIoT scenarios.
\subsubsection{Image acquisition and annotation}
\label{sec2.1.2}
Since there is a lack of publicly available datasets for chicken detection and healthy status recognition, we created our own dataset comprising a total of 30,131 images. This dataset includes 15,470 images of healthy chickens and 14,661 images of sick chickens. And the "sick chicken" category encompasses various statuses of chickens, including frailty, fearfulness, and sex stunting syndrome, characterized by small combs, etc. All images in the dataset also be categorized into 14,660 images of white-feathered chickens and 15,471 images of yellow-feathered chickens, representing the two main types of chickens found in poultry farms. These chickens are exclusively sourced from layer houses in Tianjin city, China, and fall within the age range of 20 to 60 weeks. It is noteworthy that these chickens are specifically layers bred for egg production and are not intended for consumption. Specifically, 10,138 images of both healthy and sick chickens were manually captured in a layer house using a high-quality SONY $\alpha$1 camera at a resolution of 1920 $\times$ 1080 to provide more details during model training. The remaining 19,993 images in both categories were captured during the system's daily autonomous patrolling, as shown in Fig.\ref{fig2} (a), using the intelligent camera depicted in Fig.\ref{fig2} (b), with a resolution of 320 $\times$ 320. These images were collected over a period of five months from more than 2,700 randomly chosen chickens, with each chicken being photographed multiple times on different days to ensure sample diversity.

\begin{figure*}[htbp]%
\centering
\captionsetup[subfigure]{labelformat=empty}
\begin{subfigure}
{\includegraphics[width=\textwidth]{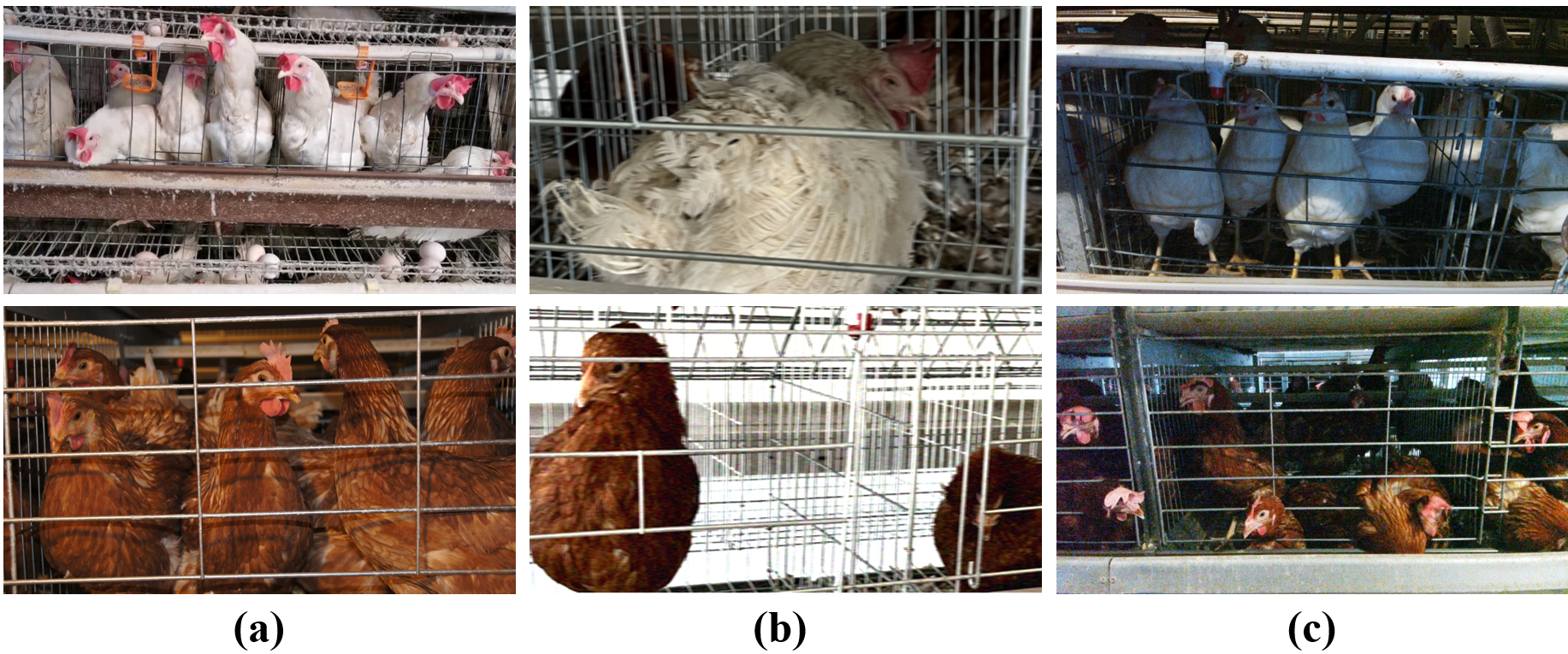}}
\end{subfigure}
\begin{subfigure}
{\includegraphics[width=\textwidth]{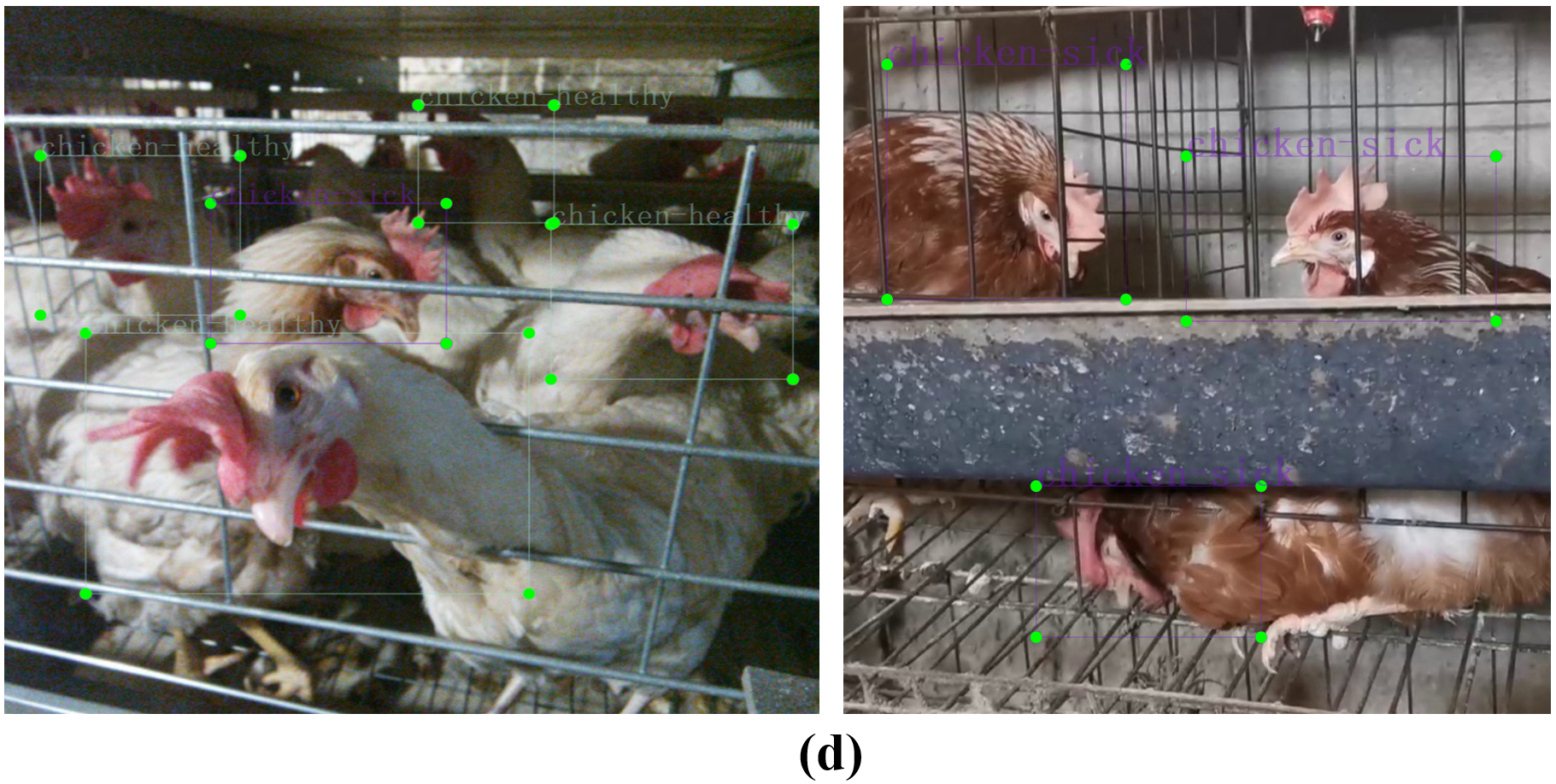}}
\end{subfigure}
\caption{An example of the training dataset: (a) and (b) show high-quality images of healthy and sick chickens, respectively, (c) and (d) display healthy and sick chickens captured from real scenarios, respectively, with annotation labels included in (d).}\label{fig20}
\end{figure*}

And all data in our dataset were manually annotated using publicly available labeling tools such as "LabelImg", under the guidance of experts in layer breeding affiliated with an academic institution of agricultural sciences. As an illustrative example, some images used for training purposes are shown in Fig.\ref{fig20}. High-quality images of healthy and sick chickens are shown in Fig.\ref{fig20} (a) and (b), respectively. Images captured from real scenarios using the intelligent cameras are shown in Fig.\ref{fig20} (c) (depicting healthy chickens) and (d) (depicting sick chickens). Additionally, Fig.\ref{fig20} (d) show labels annotated using "LabelImg", where light green boxes and purple boxes denote healthy chickens and sick chickens, respectively. It's worth noting that thanks to the well-tuned ISP of the edge-AI enabled CMOS sensor, the images captured in real scenarios (refer to Fig.\ref{fig20} (c) and (d)) maintain good quality even under the capturing conditions with movements. 

\begin{table*}[h]
\caption{Distribution details of sub-datasets.}\label{tab0}
\begin{threeparttable}
\begin{tabular*}{\textwidth}{@{\extracolsep\fill}lccccccccc}
\toprule%
 & & \multicolumn{2}{@{}c@{}}{Train}& \multicolumn{2}{@{}c@{}}{Valid}& \multicolumn{2}{@{}c@{}}{Test} & Total \\\cmidrule{3-8} %
Class &Breed  & HQ\tnote{a} & LQ & HQ & LQ & HQ & LQ & \\\midrule
 \multirow{2}{*}{Healthy} & W\tnote{b} & 2322 & 4501 & 222 & 470 & --& 195 & 7710 \\
 & Y & 2327 & 4481 & 267 & 460 & --& 225 & 7760 \\\midrule
  \multirow{2}{*}{Sick} & W & 2156 & 4080 & 244 & 440 & --& 30 & 6950 \\
 & Y & 2340 & 4552 & 260 & 509 & --& 50 & 7711 \\
\bottomrule
\end{tabular*}
\begin{tablenotes}
\footnotesize
\item[a]{"HQ" and "LQ" denote the high-quality and low-quality images which are captured by high-quality camera and intelligent cameras, respectively.}
\item[b]{"W" and "Y" short for white feathered and yellow feathered chickens, respectively.}
\end{tablenotes}
\end{threeparttable}
\end{table*}

\subsubsection{Dataset construction}
\label{sec2.1.3}
Our dataset is divided into three subsets: training, validation, and testing sub-datasets, for the purposes of training, evaluation, and implementation test respectively. The distribution details of our dataset is shown in Table.\ref{tab0}. In the table, "HQ" and "LQ" denote the high resolution images captured by high-quality camera and low resolution images captured in real scenario, respectively. And white feathered and yellow feathered chickens are represented by "W" and "Y" respectively. As shown in Table.\ref{tab0}, the testing sub-dataset comprises 500 images captured from real scenarios, while the remaining images in the training and validation sub-datasets are divided approximately in a 9:1 ratio. For each sub-dataset, we shuffled and randomly selected the images and made efforts to achieve a balanced sample distribution for both "healthy" and "sick" categories, as well as for breeding classes based on white and yellow feathered chickens, to the best of our ability. However, as shown in Fig.\ref{fig20} (d), despite the nearly equal distribution of image numbers between the "healthy" and "sick" categories, in most images of caged chickens in real scenarios, "healthy" chicken samples outnumber "sick" chicken samples in the real layer houses. Hence, addressing this sample imbalance issue will be a focus of our future work. 

\subsection{Methods}
\label{sec2.2}
\subsubsection{FCOS-Lite network structure}
\label{sec2.2.1}

\begin{figure*}[htbp]%

\centering
\makebox[\textwidth][c]{\includegraphics[width=\textwidth]{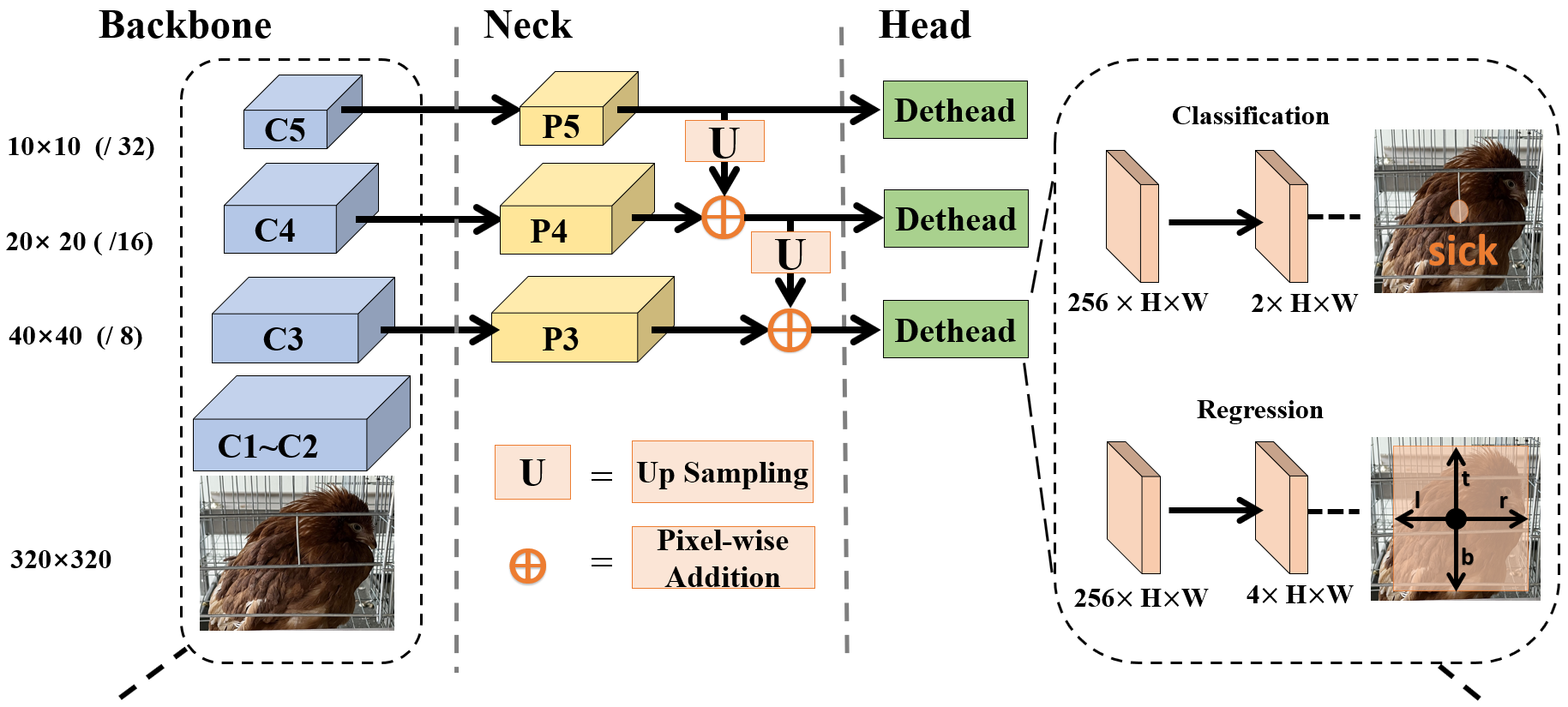}}


\centering
\makebox[\textwidth][c]{\includegraphics[width=\textwidth]{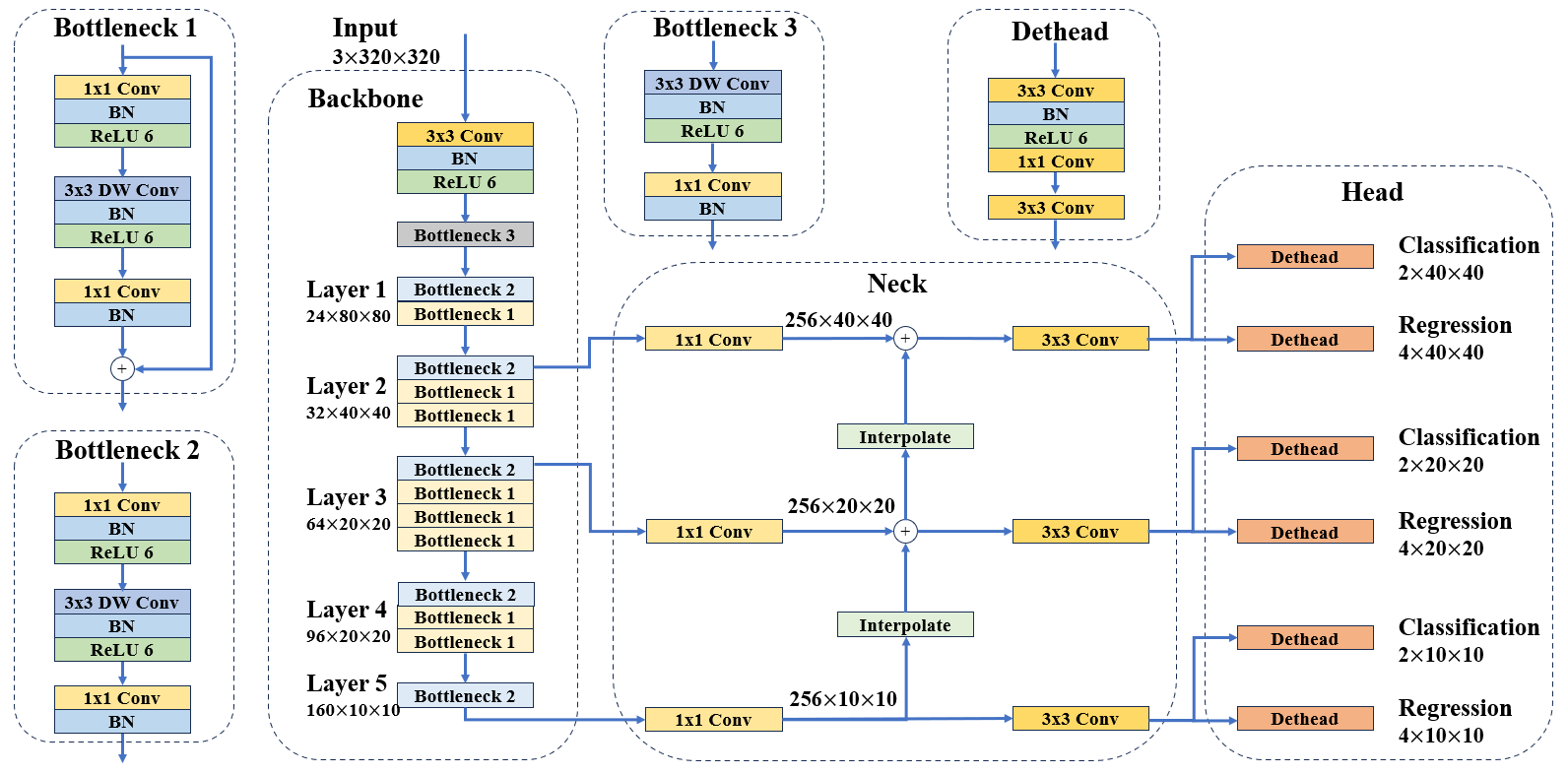}}
\caption{FCOS-Lite network structure.}\label{fig3}
\end{figure*}

To adapt the FCOS detector for edge devices, we introduce FCOS-Lite, a streamlined version optimized for lightweight processing. The schematic and detailed network structure of FCOS-Lite are illustrated in the top and bottom sections of Fig.\ref{fig3}, respectively. In comparison to the original FCOS detector, the FCOS-Lite detector include the following modifications: 
\begin{enumerate}
\item{Changing the backbone of the network from "ResNet" \cite{dl3} to "MobileNetV2" \cite{dl2}, to achieve a compact and lightweight model.}
\item{Reducing the number of FPN levels in the neck of the network from five to three, to decrease model complexity.}
\item{Modifying the components of the shared heads in the network and eliminating the original center-ness heads, to reduce model complexity.}
\end{enumerate}

Here, we only focus on introducing the components of FCOS-Lite that different from the original FCOS detector. As shown in Fig.\ref{fig3}, the dimensions of the input image are 3 channels $\times$ 320 height $\times$ 320 width. And the selection of a small input size is important to accommodate all processing tasks, including image processing, model inference, and post-processing, within the memory constraints of the edge-AI CMOS sensor. From the "MobileNetV2" backbone, three specific feature maps are chosen to produce three pyramid feature maps within the network's neck. This process is achieved by employing 1 $\times$ 1 convolutional layers with the top-down connections. And the strides of the pyramid feature maps are set at 8, 16, and 32, corresponding to the range from large to small sizes of maps, respectively. The repetitive layer sets found in the original FCOS detection heads are modified into a unified block set (referred to as "Dethead" in Fig.\ref{fig3}), consisting of a sequence of layers:  a 3 $\times$ 3 convolutional layer, batch normalization, Relu6 activation, a 1 $\times$ 1 convolutional layer and another 3 $\times$ 3 convolutional layer. These "Dethead" block sets can achieve a more compact head design and good performance based on our experimental results. Furthermore, the structure of sharing heads between different feature levels, as seen in the original FCOS, is retained in the FCOS-Lite detector for parameter efficiency. However, the center-ness head in original FCOS is omitted, since its marginal performance improvement (only 0.6 $\%$) was outweighed by its demand for an additional 1.2 $\%$ of memory space in the CMOS sensor. The classification channels for various feature levels consist of two sub-channels, corresponding to the two classes ("healthy" and "sick") of chickens. Meanwhile, the regression heads maintain four sub-channels, consistent with the original FCOS, denoting the coordinates of the bounding boxes' left (l), top (t), right (r), and bottom (b) coordinates for each instance.

\begin{table*}[h]
\caption{Size and accuracy comparison between original FCOS and FCOS-Lite.}\label{tab10}
\begin{threeparttable}
\begin{tabular*}{\textwidth}{@{\extracolsep\fill}lcccccc}
\toprule%
 Model & Params (M) & Ratio/size $\uparrow$\tnote{a} & mAP@0.5 ($\%$) & Ratio/$\Delta \downarrow$\tnote{b}  \\\midrule%
FCOS\tnote{c} & 28.4 & -- & 96.1 & -- \\
FCOS-Lite & 2.84 & 90$\%$ & 84.7 & 11.9$\%$ \\
\bottomrule
\end{tabular*}
\begin{tablenotes}
\footnotesize

\item[a]{"Ratio/size" represents the size reduction ratio of FCOS-Lite compared to the original FCOS, $\uparrow$ indicates that a higher value is better.}
\item[b]{"Ratio/$\Delta$" represents the accuracy loss ratio of FCOS-Lite compared to the original FCOS, $\downarrow$ indicates that a lower value is better.}
\item[c]{Backbone of original FCOS is ResNet50.}
\end{tablenotes}
\end{threeparttable}
\end{table*}

Finally, as shown in Table.\ref{tab10}, the "PyTorch" version (Float32) of the FCOS-Lite detector exhibits parameter count of 2.84 MB. Compared to the original FCOS (ResNet50 backbone) detector which has parameter size of 28.4 MB, the FCOS-Lite detector achieves a remarkable model size reduction ratio of 90$\%$. Following int8 quantization, the model size of FCOS-Lite can be compressed to 3.3 MB, making it sufficiently compact to accommodate the 8 MB memory constraints of the edge-AI CMOS sensor. However, FCOS-Lite also exhibits an accuracy loss issue. In this study, the accuracy loss ratio, exemplified by mAP@0.5 due to space constraints, is approximately 12$\%$, compared to the original FCOS detector.

\subsubsection{Improved loss functions}
\label{sec2.2.2}
In order to mitigate the inherent accuracy reduction in light-weighted FCOS-Lite detectors, we propose a gradient weighting loss function for classification, replacing the original Fcoal loss \cite{dl6}. Additionally, the CIOU loss function \cite{dl7} is introduced for location regression, replacing the original IoU loss \cite{dl8}. Through the implementation of these two loss functions, enhancements in the accuracy of the FCOS-Lite detector can be achieved without the need for structural modifications or incurring additional inference costs.

The Fcoal loss, utilized in the original FCOS detector, mitigates the contribution of loss from easy examples while emphasizing those from hard examples, thereby addressing the class imbalance issue. However, its performance heavily relies on hyper-parameter tuning and lacks adaptability to dynamic changes in data distribution during training. To address these limitations, we propose the gradient weighting loss. This novel approach adjusts loss weights based on the gradient norms of samples, enabling adaptation to changing data distributions and the model's learning state. By utilizing a simple threshold, the proposed gradient weighting loss assigns lower weights to easy and “outlier” samples, thereby redirecting attention to other samples.  

In the proposed gradient weighting loss, let $p\in [0,1]$ denote the probability predicted by the FCOS-Lite model for a candidate sample, and $p^{*}\in \left\{0,1\right\}$ represent its ground-truth label for a particular class. Consider the binary cross entropy loss as follow:

\begin{equation}
L_{BCE}(p,p^{*}) =  \begin{cases} -log(p) & if p^{*}=1 \\ -log(1-p) & if p^{*}=0 \end{cases} \label{eq1}
\end{equation}



Then, the norm of gradient of $p$ is denoted as $g$:

\begin{equation}
g = |p-p^{*}|= \begin{cases} 1-p & if  p^{*}=1 \\ p & if  p^{*}=0 \end{cases} \label{eq3}
\end{equation}

We then denote the loss weight $\omega$ for each sample based on its gradient norm $g$ as:

\begin{flalign}
\omega = \begin{cases} e^{g}  & if g<\mu \\ |2e^{\mu}-e^{g}| & otherwise \end{cases} \label{eq4}
\end{flalign}
where $e$ denotes exponential function, and $\mu$ represents the simple threshold for distinguishing  the “outlier” samples based on the gradient norm $g$.

And the final weighted classification loss based on binary cross entropy loss is denoted as follows:

\begin{align}
&L_{WCE}(p,p^{*},g,\mu) = \omega L_{BCE}(p,p^{*})\\\nonumber
  &= \begin{cases} e^{g}L_{BCE}(p,p^{*}) & if g<\mu \\ |2e^{\mu}-e^{g}|L_{BCE}(p,p^{*}) & otherwise \end{cases} \label{eq5}
\end{align}

\begin{figure}[htbp]
\centering
\includegraphics[width=0.5\textwidth]{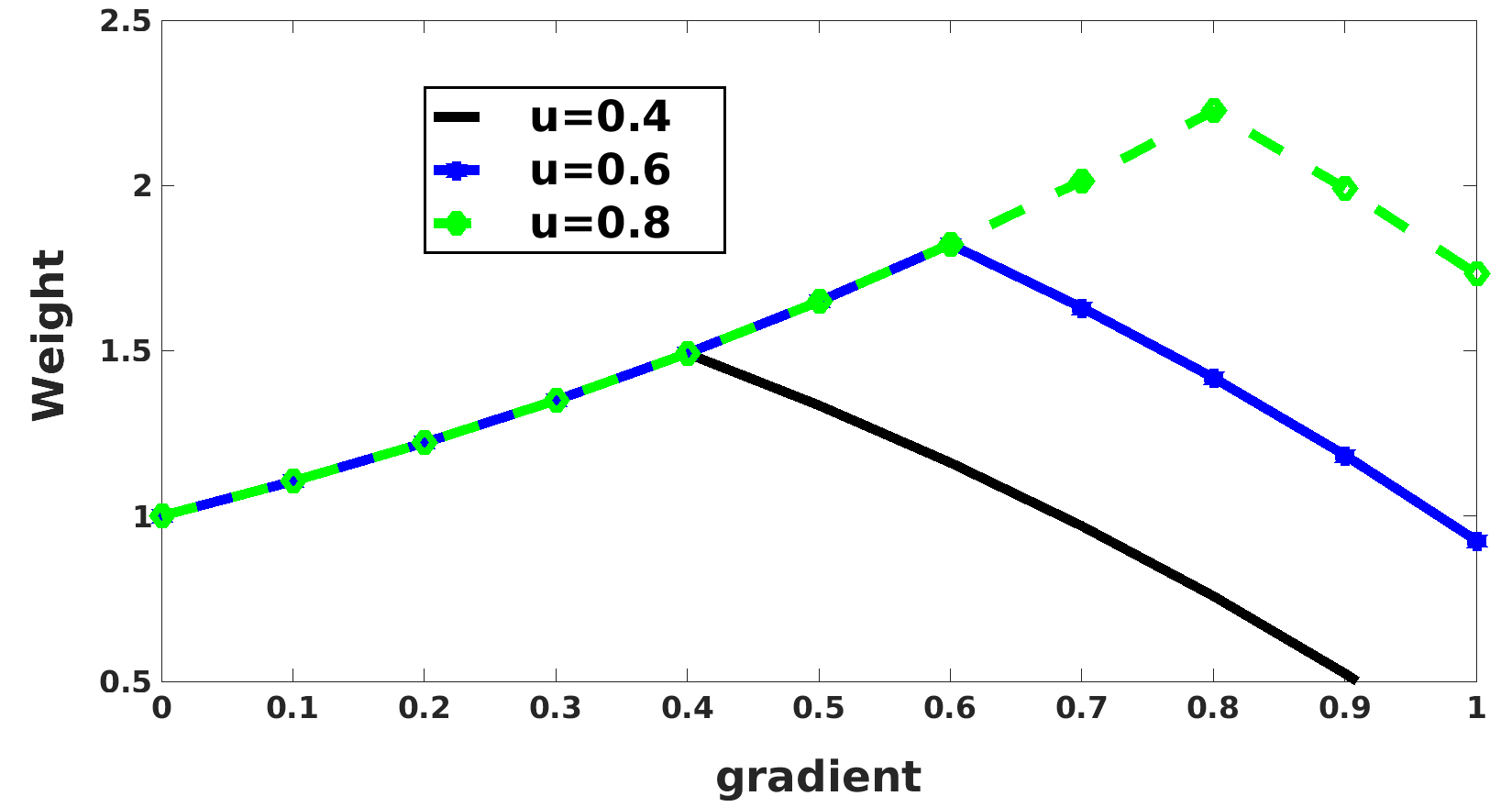}
\caption{An example of weights based on the gradient norms for classification loss, with thresholds $\mu$ are set to 0.4, 0.6, 0.8.
respectively.}
\label{fig4}
\end{figure}

As shown in Fig.\ref{fig4}, the weight $\omega$ (vertical axis) significantly increases for samples with larger gradient norms (horizontal axis) owing to the exponential function. Conversely, the weight $\omega$ decreases for "outlier” samples with gradient norms exceeding the predefined threshold $\mu$. By tuning the threshold $\mu$ (refer to $\mu$=0.4, 0.6, and 0.8 respectively in Fig.\ref{fig4}), we can adjust the range of "outlier" (very hard) samples and their contributions to the loss, thereby regulating the amount of attention paid to those outlier samples.  

\begin{figure}[htbp]%
\centering
\subfigure[]{\includegraphics[width=0.5\textwidth]{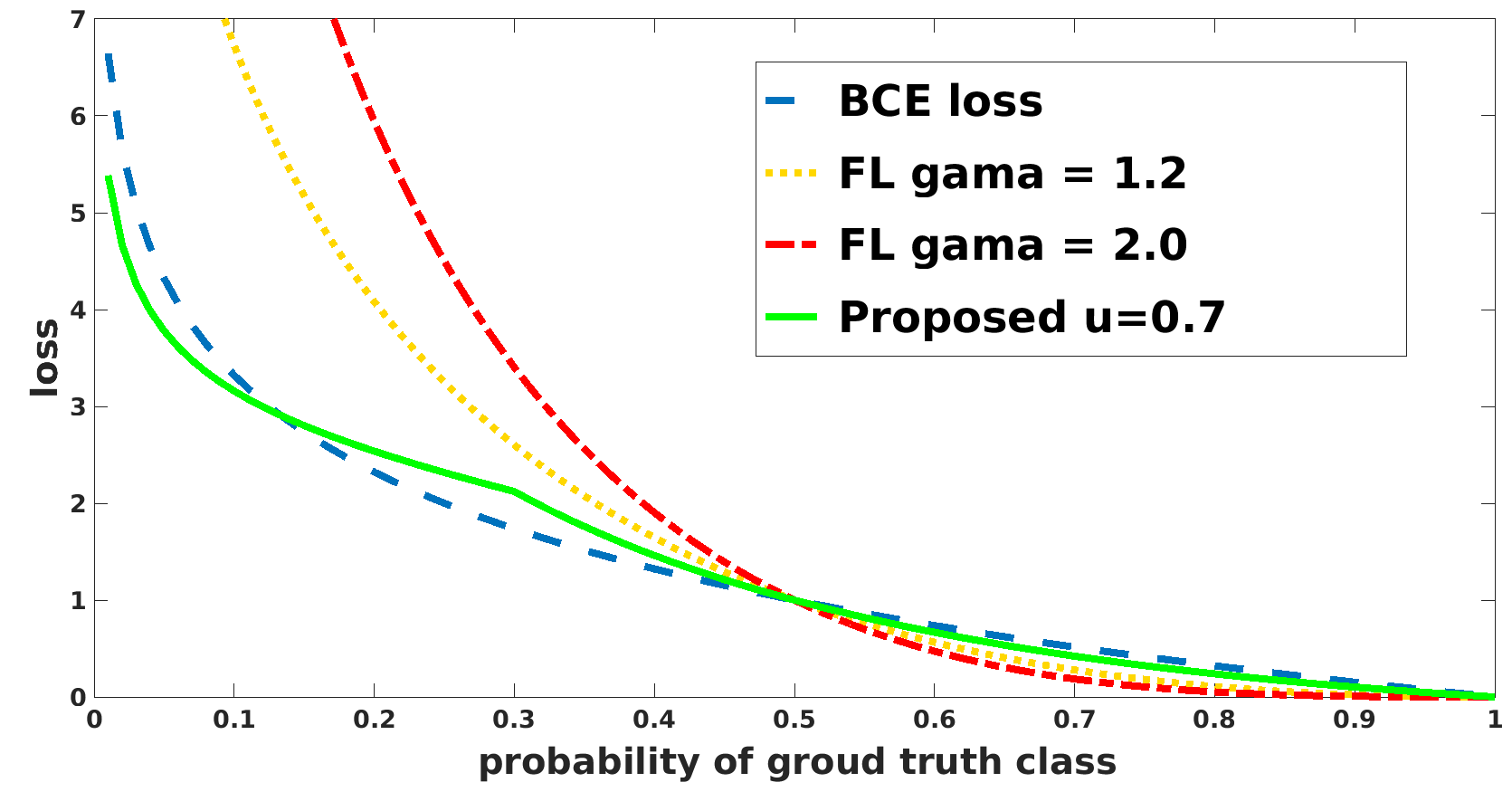}}

\vspace{-2mm}
\subfigure[]{\includegraphics[width=0.4\textwidth]{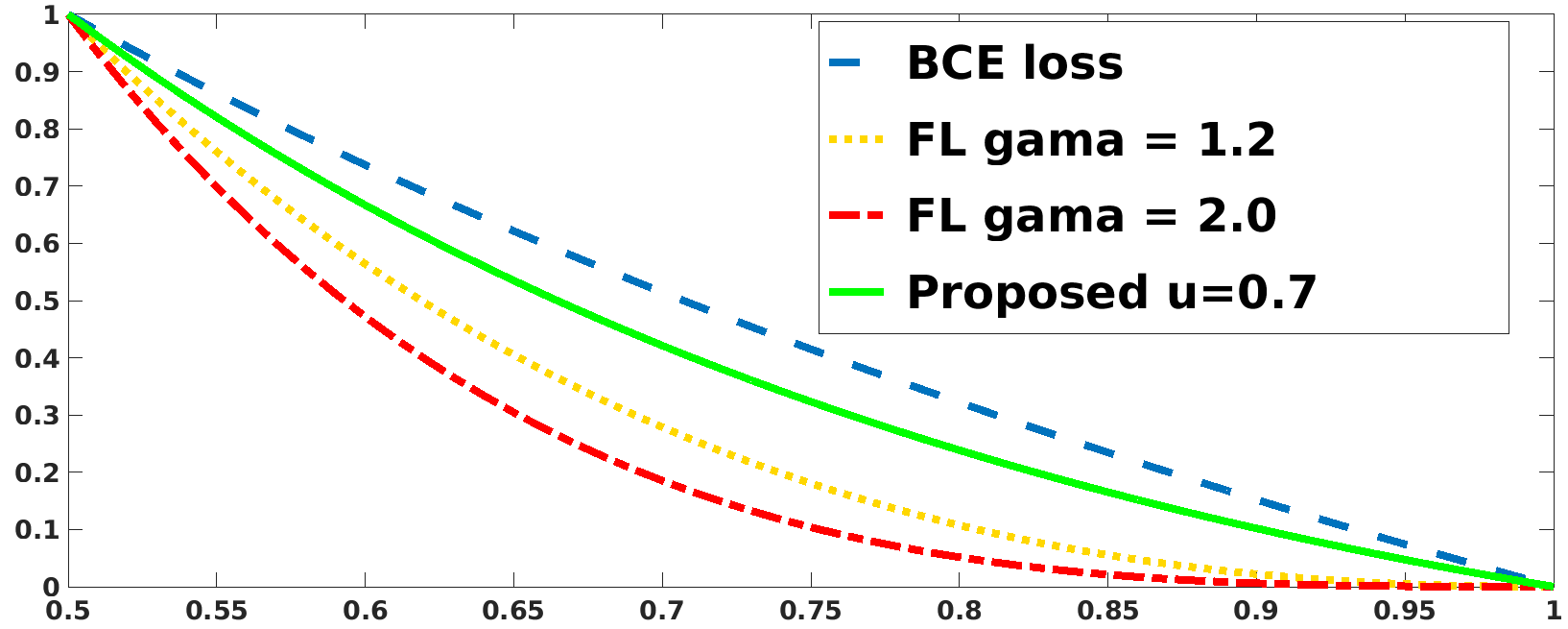}}

\caption{An example of comparison between the proposed loss function and existing loss functions, (a) all losses normalized to "1" when probability $p=0.5$, (b) zoom in perspective on the portion where probability $p>0.5$.}\label{fig5}
\end{figure}

As shown in Fig.\ref{fig5}, we present a comparison of losses between traditional binary cross-entropy (BCE) loss, Focal loss (FL) with $\gamma$ values set to 1.2 and 2.0 in the factor $(1-p)^{\gamma}$ respectively, and the proposed gradient weighting loss with $\mu$ set to 0.7. We classify samples with probability $p>0.5$ as "well-classified" samples for ground truth $p^{*}=1$, following the definition in Focal loss, and normalize the losses of all methods to "1" when $p=0.5$ for fair comparison. Compared to the Fcoal loss, which assigns excessively large loss values for outlier samples ($p$ close to zero), the proposed loss function assigns highly down-weighted loss for those outlier samples. Additionally, the loss values assigned by the proposed loss function are lower than those of BCE loss for those outlier samples. This ensures that the proposed loss function effectively mitigates the adverse effects from those outlier samples. Fig.\ref{fig5} (b) provides a "Zoom In" perspective on the portion where probability $p>0.5$. We can observe that the proposed loss function results in a decreased final classification loss for easier samples compared to the BCE loss. This encourages the model to allocate more attention to other samples. Furthermore, unlike the FCOS loss, which heavily down-weights the loss for easier samples, the proposed loss function still pays attention to these samples, albeit to a lesser extent. Consequently, the accuracy of the FCOS-Lite detector can be enhanced by utilizing the proposed gradient weighting loss.  

The Intersection over Union (IoU) loss, as applied in the original FCOS detector, may fail to accurately represent the proximity between two bounding boxes in cases where they have no intersections. To address this limitation, the Generalized IoU (GIoU) loss \cite{lo1}, Distance-IoU (DIoU) \cite{dl7} loss and Complete IoU (CIoU) loss are proposed by incorporating additional geometric factors. Especially, the additional factors in CIoU loss include the central point distance between two bounding boxes, the diagonal length of the smallest enclosing box covering these two boxes and aspect ratio component. Thereby, the CIoU loss showcases significant improvement in both convergence speed during training, and detection accuracy compared to previous loss functions. This is the main reason we employ the CIoU loss into our method for location regression. 

Given a predicted box $B$ and a target box $B^{gt}$, with their central points $b$ and $b^{gt}$ respectively, the Intersection over Union (IoU) metric and the CIoU loss are defined as follows:

\begin{equation}
IoU =  \frac{B\cap B^{gt}}{B\cup B^{gt}} \label{eq6}
\end{equation}

\begin{equation}
L_{CIoU} =  1-IoU+\frac{\rho^{2}(b,b^{gt})}{c^{2}}+\alpha\upsilon \label{eq7}
\end{equation}

where $\rho(\cdot)$ is the Euclidean distance, $c$ is the diagonal length of the smallest enclosing box covering two boxes $B$ and $B^{gt}$, $\upsilon$ measures the consistency of aspect ratio defined in Eq. (\ref{eq8}) and $\alpha$ is a positive trade-off parameter defined in Eq. (\ref{eq9}).   

\begin{equation}
\upsilon =  \frac{4}{\pi^{2}}(arctan\frac{w^{gt}}{h^{gt}}-arctan\frac{w}{h})^{2} \label{eq8}
\end{equation}

where $w$, $h$, $w^{gt}$ and $h^{gt}$ are width and height of boxes $B$ and $B^{gt}$ respectively. 

\begin{equation}
\alpha =  \frac{\upsilon}{(1-IoU)+\upsilon} \label{eq9}
\end{equation}

The final proposed detection loss function in our FCOS-Lite detector is:

\begin{equation}
L_{det} = L_{WCE}(p,p^{*},g,\mu)+L_{CIoU} \label{eq10}
\end{equation}

\subsubsection{Proposed knowledge distillation scheme}
\label{sec2.2.3}

Knowledge distillation (KD) is a technique for model compression that doesn't alter the network structure. In recent years, there has been a growing interest in applying knowledge distillation techniques to detectors. Especially, the method proposed in \cite{kd17} employs focal distillation and global distillation to encourage the student network to learn the critical pixels, channels, and pixel relations from the teacher network. This approach enables our lightweight FCOS-Lite detector to enhance its performance by leveraging valuable insights from a larger teacher detector, without damaging its compactness. 

As shown in Fig.\ref{fig1}, both focal distillation and global distillation are achieved through the computation of focal and global distillation losses, which are calculated from the Feature Pyramid Networks (FPN) of both the neck of teacher and student detectors. In focal distillation, we first utilize the ground truth bounding box to generate a binary mask $M$, scale mask $S$ for segregating the background and foreground within the feature map. Next, spatial and channel attention masks, denoted as $A^{s}$ and $A^{c}$ respectively, are calculated from teacher detector based on attention mechanisms. These masks from the teacher detector are then utilized to guide the student detector in the focal distill loss:

\begin{equation}
\label{eq11}
\begin{aligned}
L_{focal} &=\sigma \sum_{k=1}^{C}\sum_{i=1}^{H}\sum_{j=1}^{W}M_{i,j}S_{i,j}A_{i,j}^{s}A_{k}^{c}(F_{k,i,j}^{T}-F_{k,i,j}^{S})^{2} \\ &+\beta\sum_{k=1}^{C}\sum_{i=1}^{H}\sum_{j=1}^{W}\hat{M}_{i,j}\hat{S}_{i,j}A_{i,j}^{s}A_{k}^{c}(F_{k,i,j}^{T}-F_{k,i,j}^{S})^{2} \\ &+\gamma(L_{1}(A_{T}^{s},A_{S}^{s})+L_{1}(A_{T}^{c},A_{S}^{c})) 
\end{aligned}
\end{equation}
where $\sigma$, $\beta$ and $\gamma$ are hyper-parameters to balance the loss contributions between foreground, background and regularization respectively. $C$, $H$ and $W$ represent the channel, height and width of feature maps, respectively. $F^{T}$ and $F^{S}$ denote the feature maps of the teacher detector and student detector, respectively. $\hat{M}$ and $\hat{S}$ represent the inverse binary mask and inverse scale mask to preserve the background within the feature map, respectively, while $L_{1}$ denote $L_{1}$ loss.
\begin{figure}[htbp]
\centering
\includegraphics[width=0.4\textwidth]{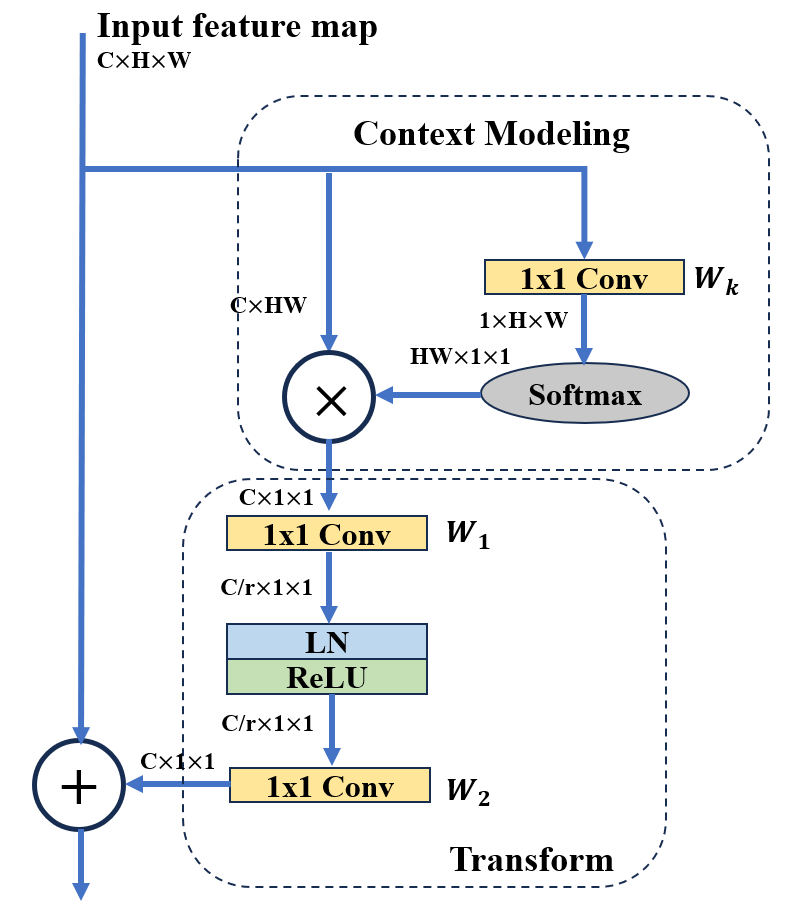}
\caption{$G_{c}Block$ employed for global distill loss calculation, its inputs are the feature maps from the necks of the teacher detector and student detector, respectively.}
\label{fig7}
\end{figure}

$G_{c}Block$ as shown in Fig.\ref{fig7} is employed to respectively capture the global relation information from the feature maps of the teacher detector and student detector. Next, the global relations from the teacher detector guide the student detector using the global distillation loss:  

\begin{equation}
L_{global} = \lambda \sum(G(F^{T})-G(F^{S}))^{2} \label{eq12}
\end{equation}
inside
\begin{align}
G(F) = F+W_{2}(ReLU(LN(\\\nonumber
W_{1}(\sum_{j=1}^{N_{p}}\frac{e^{W_{k}F_{j}}}{\sum_{m=1}^{N_{p}} e^{W_{k}F_{m}}}F_{j}  )))) \label{eq13}
\end{align}

where $\lambda$ denote a hyper-parameter, $W_{k}(\cdot)$, $W_{1}(\cdot)$, $W_{2}(\cdot)$, $ReLU(\cdot)$ and $LN(\cdot)$ represent the outputs of convolutional layers $W_{k}$, $W_{1}$, $W_{2}$, ReLU, and layer normalization, respectively. $N_{p}$ denote the number of pixels in the feature. 

Finally, based on Eq. (\ref{eq10}) to (\ref{eq12}), the overall training loss function for our FCOS-Lite detector within the knowledge distillation scheme is as follows: 
\begin{equation}
L = L_{focal}+L_{global}+L_{det} \label{eq14}
\end{equation} 

\subsection{Model training}
\label{sec2.3}

We implemented the proposed detector in PyTorch for training. The hardware configuration comprised an Intel Xeon Silver 4214R CPU with 24 cores, operating at a frequency of 2.40 GHz per core. The system is equipped with 256 GB of memory and utilizes an NVIDIA RTX 3090 Ti GPU with 24 GB of GDDR6X memory. The operating system version is Ubuntu 18.04, while the versions of Python, PyTorch, CUDA, and cuDNN are 3.8, 1.11.0, 11.3, and 8.2, respectively.

During training, all 26,759 images from the training sub-dataset were used. The input image size and batch size were configured as $320\times320\times3$ (height $\times$ width $\times$ channel) and 32, respectively. The model training for the teacher model, student model, and knowledge-distilled model took 40, 40, and 50 epochs (836 iterations per epoch), respectively. The final model was selected based on the best total accuracy in terms of "mAP" and "F1-score on validation sub-dataset. We employed SGD as the optimizer, with initial and final learning rates set to 2e-3 and 2e-5, respectively. The learning rates were reduced by a factor at iteration 24,000 and 28,000, respectively. Additionally, weight decay and momentum were set to 1e-4 and 0.9, respectively. The IoU threshold for Non-Maximum Suppression (NMS) was set to 0.6. As for the hyper-parameters of knowledge distillation, they were configured as follows: $\sigma=1.6\times10^{-3}$, $\beta=8\times10^{-4}$, $\gamma=8\times10^{-4}$, $\lambda=8\times10^{-6}$ and temperature $t=0.8$. 

\subsection{Model evaluation and deployment}
\label{sec2.4}

To coherently and fairly evaluate and compare the performance of our proposed detector, we used the PyTorch platform and assessed the detector based on the following indicators: mAP@0.5 (mean Average Precision with Intersection over Union threshold set to 0.5), Precision (P), Recall (R), F1-score and Specificity for class recognition. These metrics were calculated as follows:

\begin{equation}
P = \frac{TP}{TP+FP} \label{eq15}
\end{equation}

\begin{equation}
R = \frac{TP}{TP+FN} \label{eq16}
\end{equation}

\begin{equation}
F_{1} = \frac{2PR}{P+R} \label{eq17}
\end{equation}

\begin{equation}
mAP = \frac{\sum_{1}^{N}AP}{N}=\frac{\sum_{1}^{N}\int_{0}^{1}P(R)dR}{N} \label{eq18}
\end{equation}

\begin{equation}
Specificity = \frac{TN}{TN+FP} \label{eq21}
\end{equation}

where $TP$ (true positive) represents the count of samples accurately classified by the detector into their respective status categories (healthy or sick). Conversely, $FP$ (false positive) denotes the instances incorrectly classified by the detector as belonging to a status category when they do not. Similarly, $FN$ (false negative) refers to the count of samples erroneously categorized into the opposite status category. $AP$ corresponds to the area under the precision-recall curve, while $mAP$ signifies the average precision across different categories. $N$ is assigned a value of 2, representing the total number of categories being evaluated.   

Additional, model parameters and GFLOPs (Giga Floating Point Operations Per Second) were used to measure the computational efficiency and memory requirements of the models. Moreover, to verify the deployability and performance of the model on a memory-constrained edge-AI enabled CMOS sensor (with total memory of 8 MB and actual memory requirement for possible running being less than 5 MB), the models for comparison were all converted to TF-Lite versions with int8 quantization for measuring their actual model sizes. Finally, we implemented the TF-Lite version of the proposed edge-AI enabled detector on the CMOS sensor and verified its performance, including accuracy and inference speed measured in FPS.

\section{Experimental results}
\label{sec3}
\subsection{Evaluation of model improvements}
\label{sec3.2}
This section examines the influence of the proposed modifications implemented in our detector using our own dataset. To ensure a fair comparison, all methods are implemented in PyTorch.

\begin{table*}[h]
\caption{Ablation study results of the proposed FCOS-Lite detector with variant loss functions.}\label{tab1}
\begin{threeparttable}
\begin{tabular*}{\textwidth}{@{\extracolsep\fill}lcccccccccccccc}
\toprule%
 \multicolumn{2}{@{}c@{}}{Loss\tnote{a}} & mAP@0.5  & AP ($\%$) & P  & R  & F1 \\\cmidrule{1-2} %
 reg  & cls & ($\%$) & (healthy/sick) & ($\%$) & ($\%$) & ($\%$)\\
\midrule
 \multirow{3}{*}{IoU} & FL\tnote{b} & \textbf{84.7} & 74 / 95.4 & 83.4 & 84 & \textbf{83.7} \\
 & BCE & 85.6 & 80.2 / 91 & 84.6 & 86 & 85.3 \\
  & WCE & \textbf{88.4} (+3.7) & 84.2 / 92.6 & 87.4 & 88.2 & \textbf{87.8} (+4.1) \\ \midrule
 \multirow{3}{*}{GIoU} & FL & 85.7 & 81.5 / 89.3 & 84.4 & 87 & 85.7 \\
  & BCE & 87.1 & 82.5 / 91.7 & 84.6 & 88.1 & 86.3 \\
  & WCE & 88.8 & 84.5 / 93.1 & 86.2 & 89 & 87.6 \\ \midrule
 \multirow{3}{*}{DIoU} & FL & 85.9 & 81.8 / 90 & 86.1 & 83.7 & 84.9 \\
  & BCE & 86.7 & 82.9 / 90.5 & 87.4 & 86.2 & 86.8 \\
  & WCE & 88.9 & 85.5 / 92.3 & 89.2 & 88 & 88.6 \\ \midrule
 \multirow{3}{*}{CIoU} & FL & 87.1 & 78.5 / 94.3 & 86.8 & 85.6 & 85.4 \\
  & BCE & 87.5 & 83.4 / 91.6 & 87.1 & 87.4 & 87.1 \\
  & WCE & \textbf{90} (+5.3) & 85.5 / 94.5 & 88.5 & 89.9 & \textbf{89.2} (+5.5) \\ 
\bottomrule
\end{tabular*}
\begin{tablenotes}
\footnotesize
\item[a]{"reg" and "cls" denote loss functions for bounding box localization and classification in the proposed FCOS-Lite detector, respectively.}
\item[b]{this is the baseline method.}
\end{tablenotes}
\end{threeparttable}
\end{table*} 

As shown in Table.\ref{tab1}, in our FCOS-Lite detector, "reg" and "cls" represent the loss functions for bounding box localization and classification, respectively. "FL", "BCE" and "WCE" refer to Focal loss, binary cross-entropy loss and the proposed gradient weighting loss, respectively. It's important to note that for optimal performance with each loss function, we fine-tuned the parameters $\alpha_{t}$ and $\gamma$ of Focal loss to 0.4 and 1.2 respectively, and the parameter $\mu$ of the gradient weighting loss to 0.7. The baseline for this ablation study comprises the combination of Focal Loss (FL) and IoU loss, which are the loss functions utilized in the original FCOS detector. The results demonstrate that the integration of gradient weighting loss and CIoU loss significantly enhances the detector's performance. Compared to the baseline method, the mAP@0.5 and F1-score show improvements of 5.3$\%$ and 5.5$\%$, respectively. Notably, gradient weighting loss proves more effective for our detector than FL and BCE losses, leading to approximately a 4$\%$ improvement in both mAP@0.5 and F1-score, compared to the baseline method. Furthermore, the results indicate that the performance achieved with GIoU and DIoU losses is comparable, but CIoU loss demonstrates superior performance for bounding box localization. Finally, we utilize the proposed FCOS-Lite detector with "WCE" loss and CIoU loss as a student detector for further knowledge distillation. 

\begin{figure}[htbp]%
\centering
\begin{minipage}[t]{\linewidth}
\includegraphics[width=\textwidth]{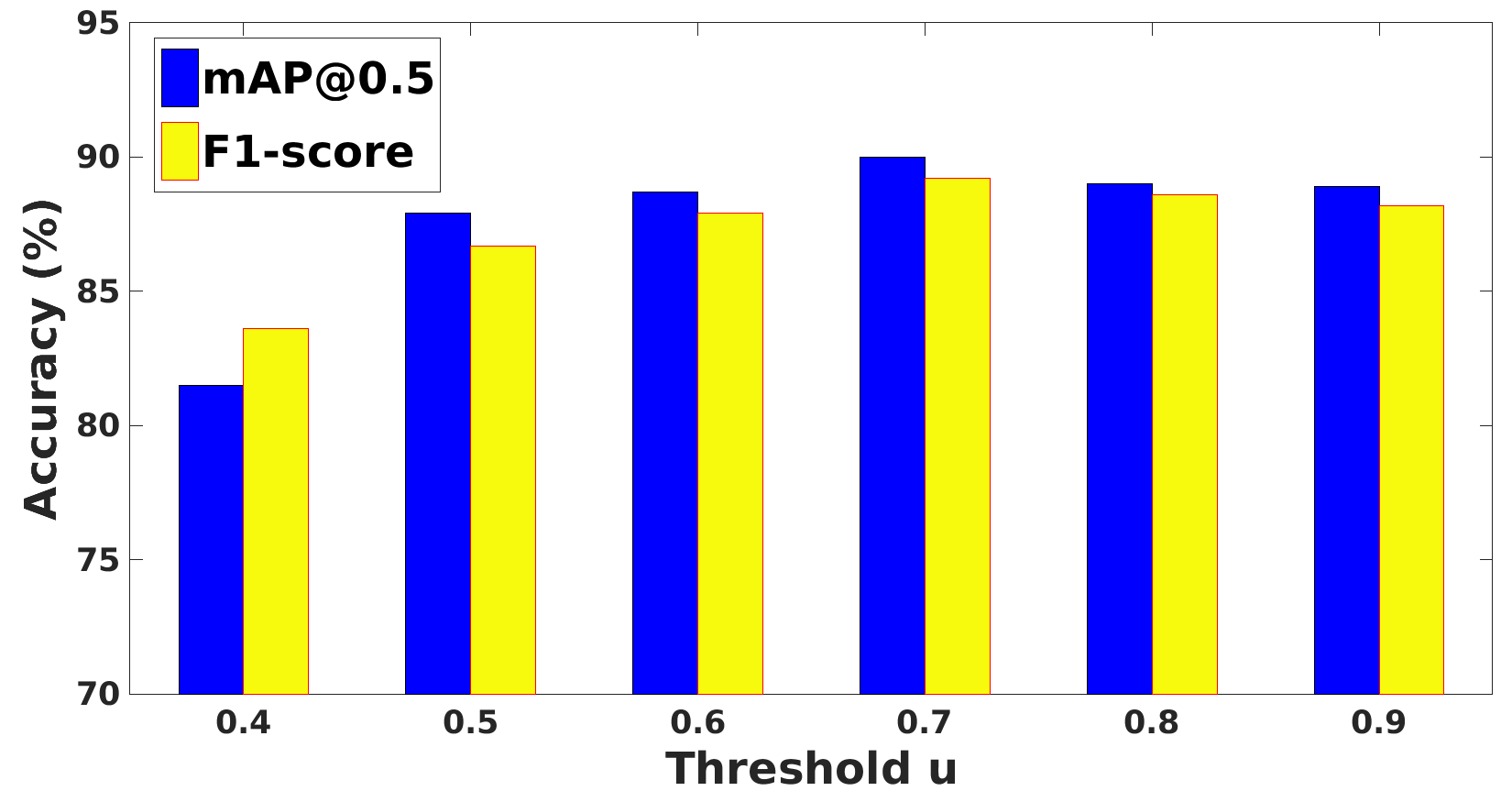}
\centering{(a)}
\end{minipage}

\begin{minipage}[t]{\linewidth}
\includegraphics[width=\textwidth]{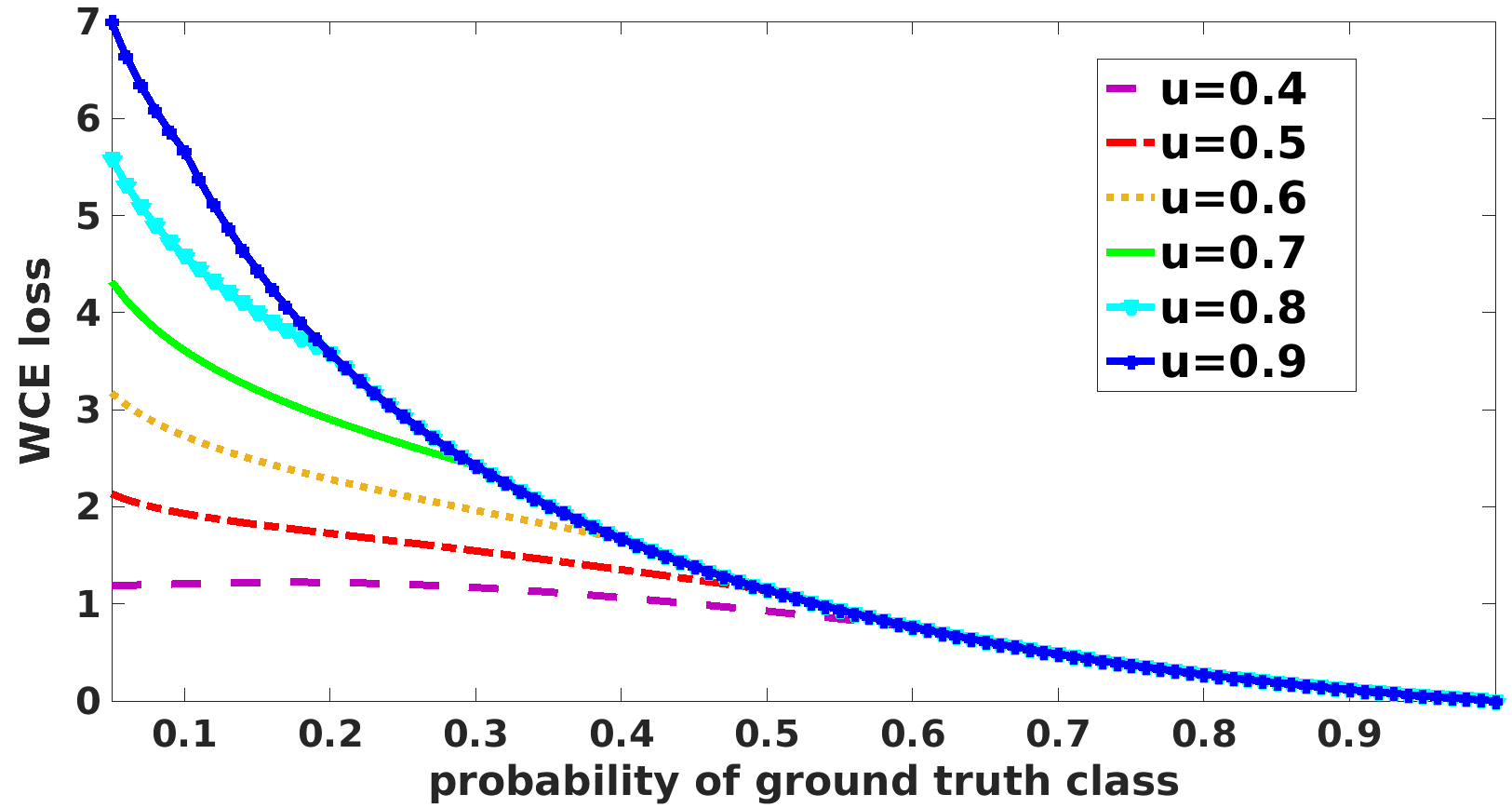}
\centering{(b)}
\end{minipage}
\caption{Comparison of accuracy and loss metrics using different threshold values $\mu$ for the proposed gradient weighting loss, (a) shows  the results of mAP@0.5 and F1-score across varying threshold values $\mu$, (b) shows the corresponding loss values for different threshold values $\mu$.}\label{fig9}
\end{figure}

Fig.\ref{fig9} shows the comparison of accuracy and loss metrics across various threshold values $\mu$ for the proposed gradient weighting loss. To accommodate space limitations, we only present mAP@0.5 and F1-score as accuracy metrics. In Fig.\ref{fig9} (a), it shows that  our proposed gradient weighting loss achieves optimal accuracy when the threshold $\mu$ is set to 0.7. Deviating from threshold $\mu$ = 0.7 results in decreased accuracy and setting the threshold $\mu$ to 0.4 leads to a significant decrease in final accuracy. That is because at $\mu$ smaller than 0.4, the gradient weighting loss values exhibit an irregular pattern (refer to Fig.\ref{fig9} (b)), failing to appropriately emphasize hard samples during training. In this study, employing a strategy of assigning lower loss values to those samples with a gradient norm of predicted class probability greater than 0.7, designated as "outlier" samples, proves to be a more effective approach for significantly improving the final accuracy. 

\begin{table*}[h]
\caption{Ablation study results of knowledge distillation.}\label{tab2}
\begin{threeparttable}

\begin{tabular*}{\textwidth}{@{\extracolsep\fill}lcccccccc}
\toprule%
T\tnote{a}& Back & Params & FLOPs & mAP@0.5  & AP-sick\tnote{b}  & P  & F1 \\ %
S& bone& (M)  & (B) & ($\%$) & ($\%$) & ($\%$) & ($\%$) \\
\midrule
B\tnote{c}& mbv2\tnote{d}& 2.84 & 1.95 & \textbf{90} & 94.5 & 88.5 &  \textbf{89.2}  \\\midrule
T& Res34\tnote{e}& 25.6 & 15.2 & 93  & 96.7 & 91.1 & 91.2 \\ 
S& mbv2& 2.84 & 1.95 & 91.3 & 95.9 & 90.7 &  90  \\\midrule
T& Res50& 28.4 & 16.3 & 96.1  & 98.9   & 95.4 & 95.2   \\
S& mbv2& 2.84 & 1.95 & \textbf{95.1} & 98.1 & 94.3 &  \textbf{94.2}  \\\midrule
T& Res101& 46.5 & 23.4 & 96.3 & 99.1  & 95.7 & 94.8 \\
S& mbv2& 2.84 & 1.95 & 95.3 & 98.7 & 94.4 &  94.0 \\
\bottomrule
\end{tabular*}
\begin{tablenotes}
\item[a]{"T" and "S" denote the results of the teacher detector and the student detector, respectively.}
\item[b]{"AP-sick" denotes the AP values for the "sick" category.}
\item[c]{"B" denotes the baseline method before knowledge distillation.}
\item[d]{"mbv2" denotes the proposed FCOS-Lite detector with MobilenetV2 backbone, which serves as the student detector.}
\item[e]{"Res*" denotes the original FCOS detector with a ResNet* backbone, which is used as the teacher detector}
\end{tablenotes}
\end{threeparttable}
\end{table*}

Table.\ref{tab2} shows the results of student detector "distilled" by different teacher detectors in our study. In this table, "mbv2" backbone refers to the proposed FCOS-Lite detector configured with a MobilenetV2 backbone and utilizing both "WCE" and CIoU losses, functioning as the student detector. On the other hand, the backbones labeled as "Res34", "Res50" and "Res101" represent the original FCOS detector employing ResNet34, ResNet50 and ResNet101 backbones, respectively, functioning as teacher detectors during the knowledge distillation process. It's crucial to highlight that we applied "WCE" and CIoU loss functions on the teacher detectors and meticulously fine-tuned the parameters of knowledge distillation for optimal performance. Moreover, due to space constraints, we only present the results for mAP@0.5, AP of the "sick" category, precision (P), and F1-score. The results for the AP of the "healthy" category and recall (R) can be derived from the presented data. As shown in Table.\ref{tab2}, when compared to the original FCOS detectors across various backbone architectures, ranging from the smallest ResNet34 to the largest ResNet101, FCOS-Lite exhibits a reduction in parameter size (approximately 1/9 $\sim$ 1/16) and computational complexity (approximately 1/7 $\sim$ 1/12 GFLOPs). However, the performance of FCOS-Lite is "compromised" by its reduced parameter size and lower computational complexity. Despite the performance enhancements achieved through the utilization of "WCE" loss and CIoU loss in the FCOS-Lite model, its overall performance is still "compromised". As we can see from Table.\ref{tab2}, after knowledge distillation (KD), the mAP@0.5 and F1-score of the student detector show improvements of at least $1.3\%$ and $0.8\%$, respectively, when the teacher detector is the FCOS detector with the ResNet34 backbone. Furthermore, these metrics experience enhancements of $5.3\%$ and $4.8\%$, respectively, when using the FCOS detector with ResNet101 backbone. However, based on the experimental results, we think that FCOS detector with ResNet50 backbone serves as the most efficient teacher model in this study. It contributes to notable improvements in the mAP@0.5 and F1-score of the student FCOS-Lite detector, enhancing them by approximately $5.1\%$ and $5.0\%$, respectively, while requiring much fewer (2/3) parameters compared to the teacher detector with ResNet101 backbone. This results in shorter training time and less resources costs. Finally, we utilize the proposed FCOS-Lite detector, which is knowledge-distilled from the teacher detector with a ResNet50 backbone, to compare it with other classic detectors with lightweight capabilities.

\subsection{Evaluation of model classification performance}
\label{sec3.11}

\begin{figure*}[htbp]%
\makebox[\textwidth][c]{
\begin{minipage}[t]{0.5\linewidth}
\centering
\includegraphics[width=\textwidth]{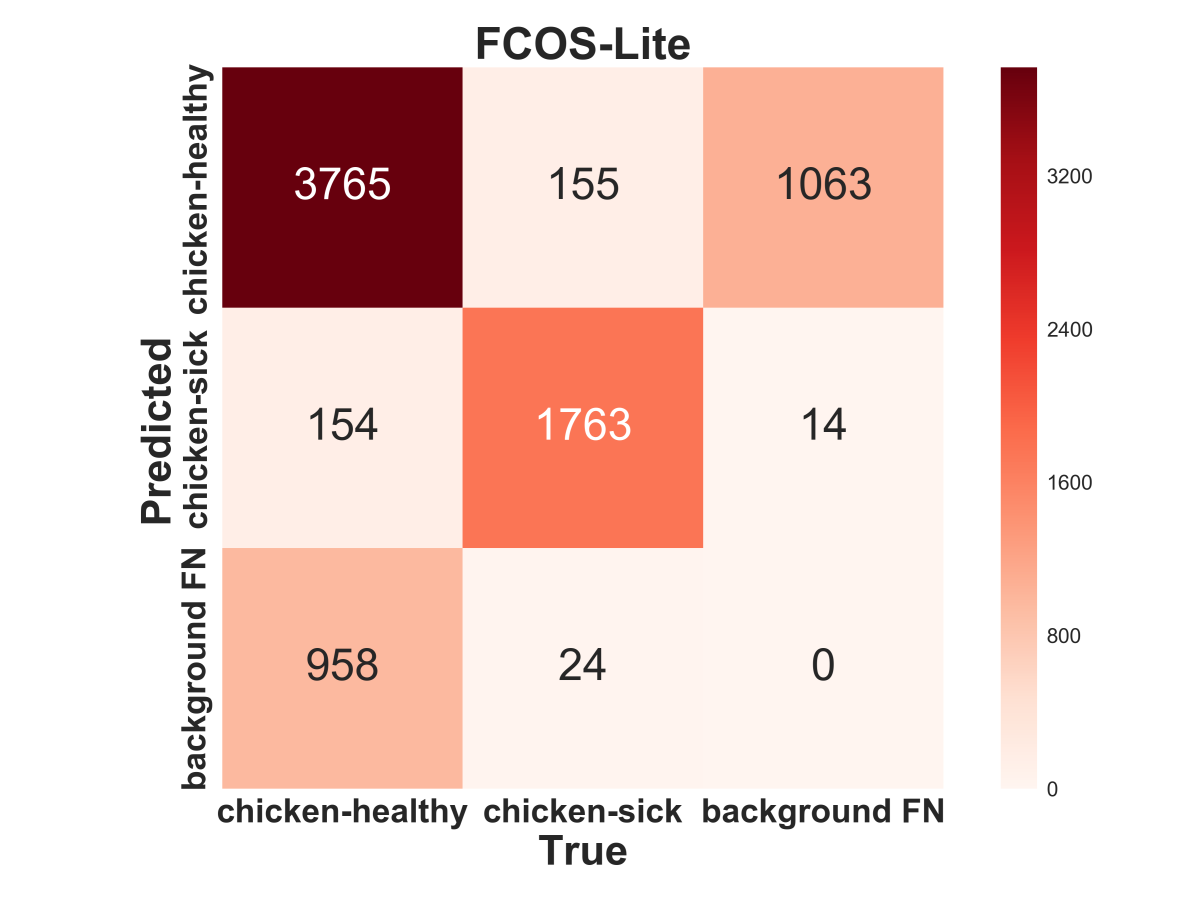}
\centering{(a)}
\end{minipage}

\begin{minipage}[t]{0.5\linewidth}
\centering
\includegraphics[width=\textwidth]{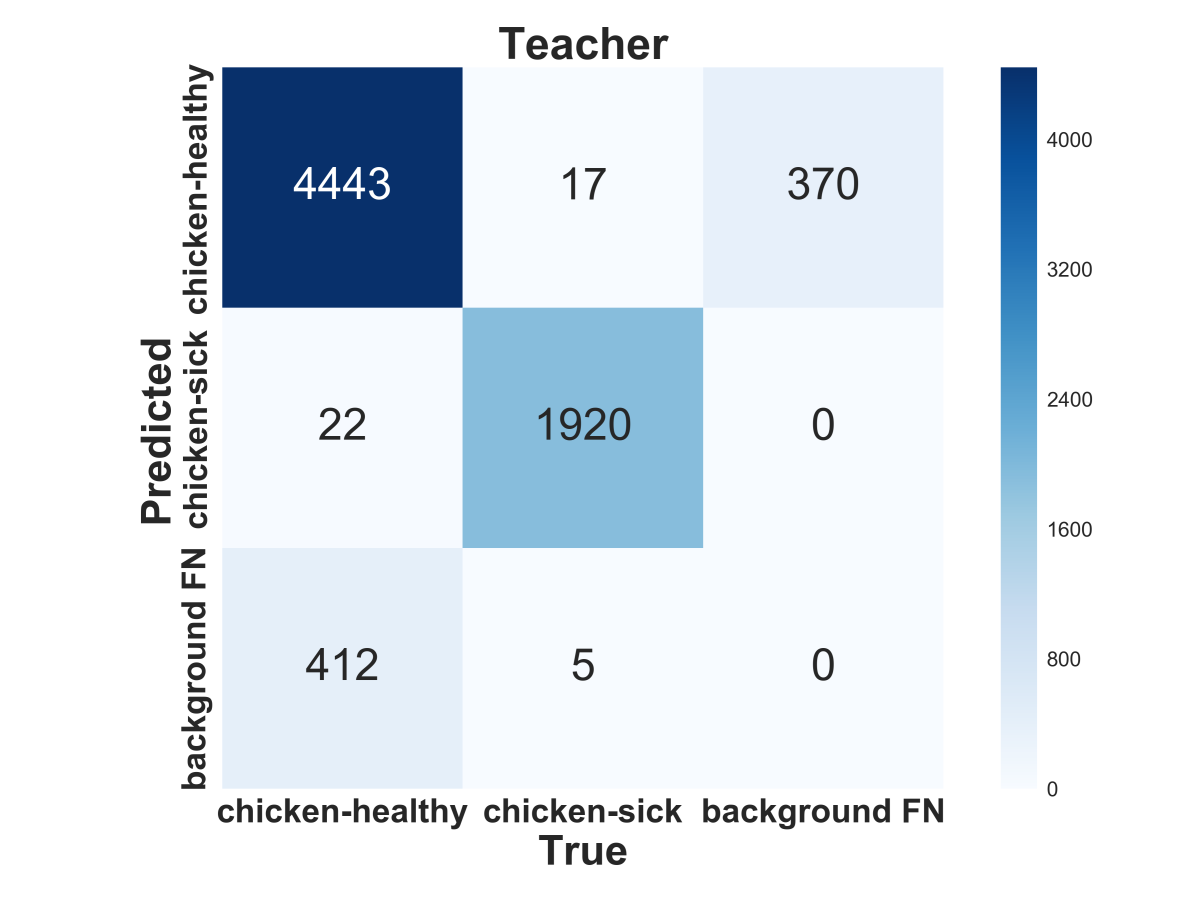}
\centering{(b)}
\end{minipage}
}
\makebox[\textwidth][c]{
\begin{minipage}[t]{0.5\linewidth}
\centering
\includegraphics[width=\textwidth]{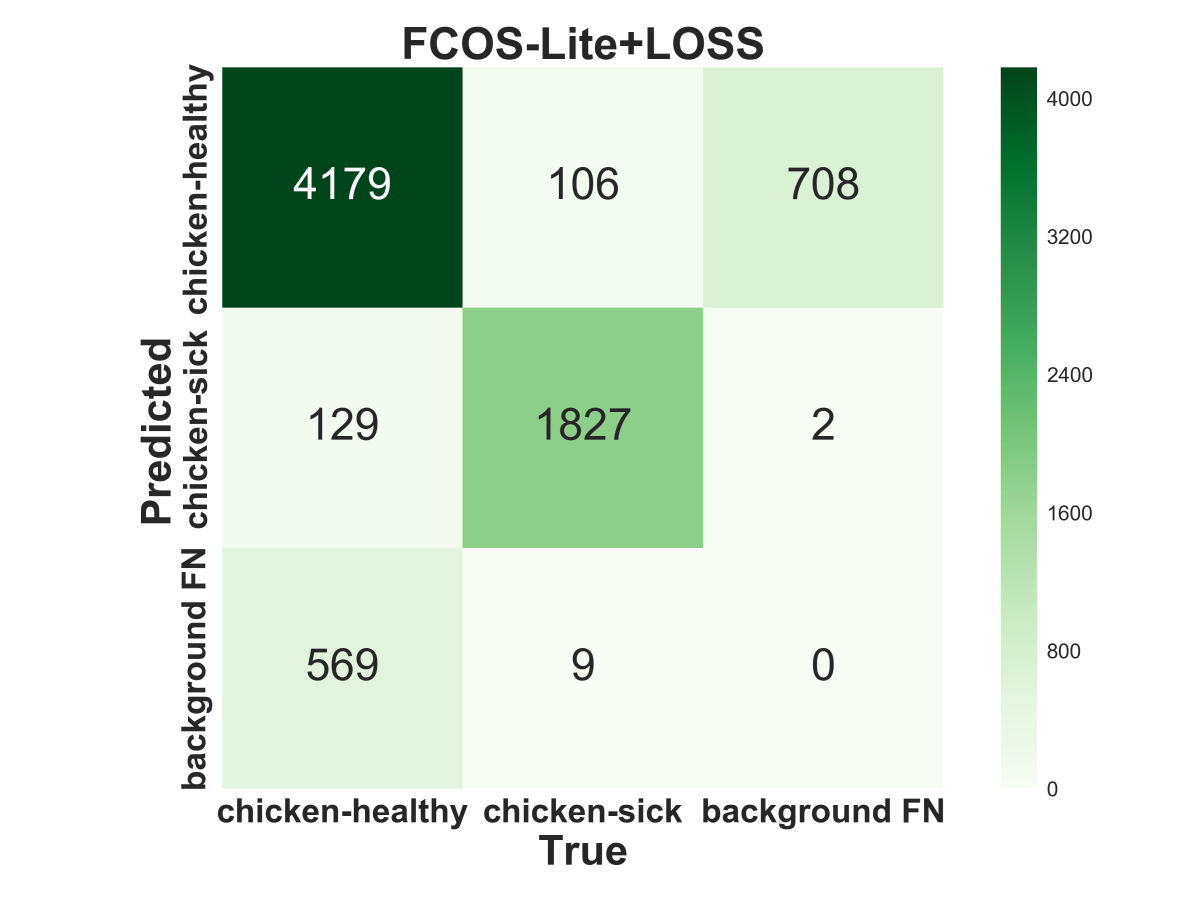}
\centering{(c)}
\end{minipage}
\begin{minipage}[t]{0.5\linewidth}
\centering
\includegraphics[width=\textwidth]{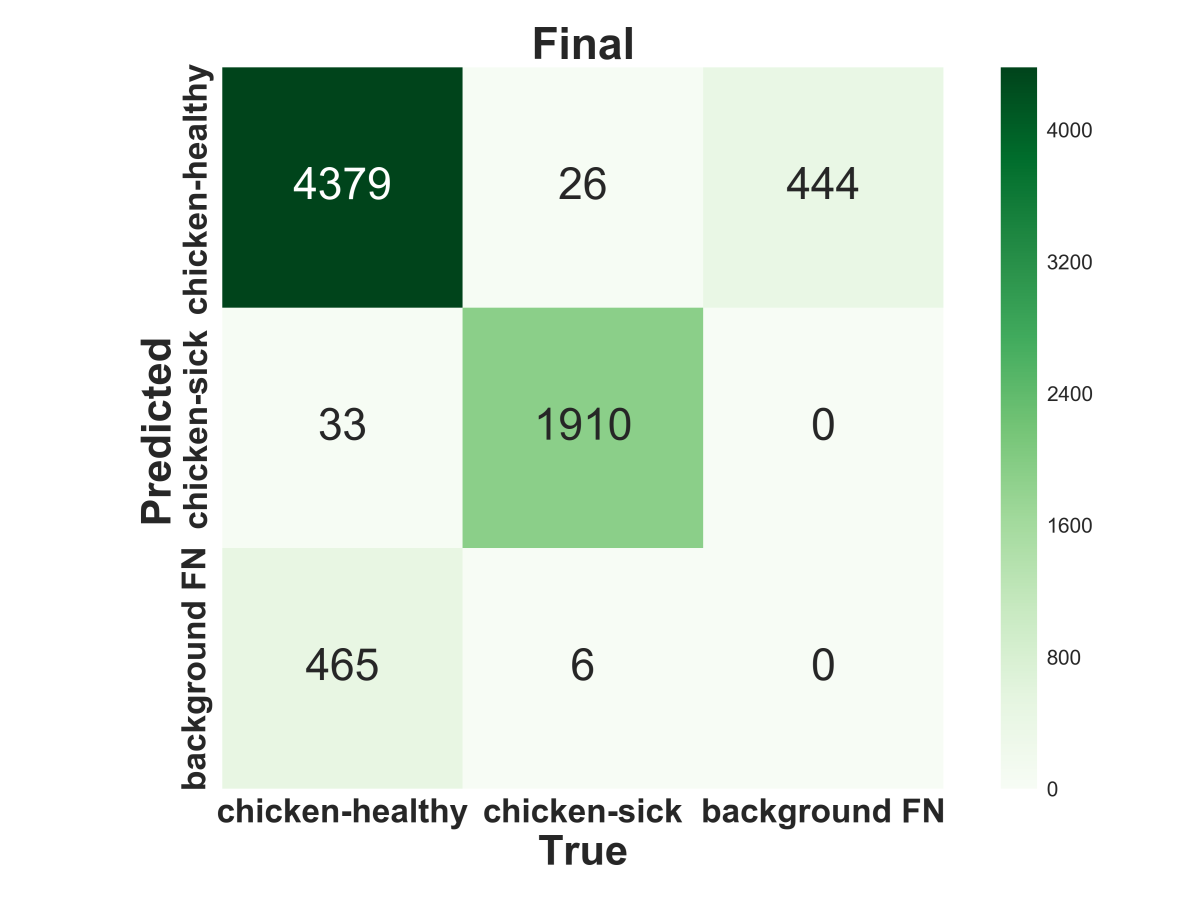}
\centering{(d)}
\end{minipage}

}

\caption{Confusion matrices for: (a) FCOS-Lite model, (b) Teacher model with ResNet50 backbone, (c) FCOS-Lite with improved loss function, and (d) Final student model after knowledge distillation.}\label{figa2}
\end{figure*}

\begin{table*}[h]
\caption{\hl{Key metrics derived from confusion matrices of models.}}\label{tabconfuse}
\begin{threeparttable}
\begin{tabular*}{\textwidth}{@{\extracolsep\fill}lccccc}
\toprule%
Model & Category &Precision (\%) & Recall (\%) & Specificity (\%)  \\ %
\midrule
\multirow{2}{*}{FCOS-Lite} & healthy & 75.6  & 77.2  & 59.7  \\
& sick & 91.3  &  90.8 & 97.2  \\\midrule
\multirow{2}{*}{+ Loss\tnote{a}} & healthy & 83.7  &  85.7 &  69.3 \\
& sick &  93.3 & 94.1  &  97.7 \\\midrule
\multirow{2}{*}{Teacher\tnote{b}} & healthy & 92.0  &  91.1 & 83.3  \\
& sick & 98.9  &  98.9 &  99.6 \\\midrule
\multirow{2}{*}{Final\tnote{c}} & healthy &  \textbf{90.3} &  \textbf{89.8} &  \textbf{80.3} \\
& sick &  \textbf{98.3} &  \textbf{98.4} &  \textbf{99.4} \\
\bottomrule
\end{tabular*}
\begin{tablenotes}
\item[a]{FCOS-Lite with improved loss function.}
\item[b]{FCOS with ResNet50 backbone as a teacher model.}
\item[c]{FCOS-Lite with improved loss function and knowledge distillation.}
\end{tablenotes}
\end{threeparttable}
\end{table*}

Fig.\ref{figa2} presents the confusion matrices of the models. The dataset comprises 4877 healthy chicken targets across 1419 images and 1942 sick chicken targets across 1453 images. As shown in Fig.\ref{figa2}, the proposed loss functions (c) and knowledge distillation scheme (d) effectively enhance the true positive rates for both healthy and sick chicken categories compared to the original FCOS-Lite model (a). Furthermore, the proposed methods effectively reduce the number of both mistaken detections and missed detections that are incorrectly classified as background (false negatives), which should ideally be close to zero. The precision, recall, and specificity percentages for each "healthy" and "sick" category, derived from the confusion matrices, are presented in Table.\ref{tabconfuse}. It is evident that the proposed methods significantly enhance classification accuracy, including "specificity," which measures the ability of the model to correctly identify negative cases.

\begin{table*}[h]
\caption{Detailed AP of sick chickens.}\label{tabsick}
\begin{threeparttable}
\begin{tabular*}{\textwidth}{@{\extracolsep\fill}lc|c|c|c|c|c|c|cc}
\toprule%
Model & \multicolumn{2}{c}{Sick Total} & \multicolumn{2}{c}{frailty} & \multicolumn{2}{c}{fear}  & \multicolumn{2}{c}{sex stunting} \\ %
 &Num.\tnote{a} & AP\tnote{b}  & Num. & AP & Num. & AP & Num. & AP \\
\midrule
 FCOS-Lite& \multirow{4}{*}{1942} & 95.4 & \multirow{4}{*}{697} & 94.9 &  \multirow{4}{*}{814} & 94.9 & \multirow{4}{*}{431} &97.2 \\ 
 + Loss\tnote{c} & & 94.5 &   & 94.1&  & 94.1& &95.9   \\ 
Teacher\tnote{d}&  & 98.9 &  & 98.8 & & 98.7 &  &99.5  \\
 Final\tnote{e}&  & \textbf{98.1} &  & \textbf{97.8} &  & \textbf{97.7} & & \textbf{99.2}\\
\bottomrule
\end{tabular*}
\begin{tablenotes}
\item[a]{number of target objects in dataset.}
\item[b]{percentage value of AP.}
\item[c]{FCOS-Lite with improved loss function.}
\item[d]{FCOS with ResNet50 backbone as a teacher model.}
\item[e]{FCOS-Lite with improved loss function and knowledge distillation.}
\end{tablenotes}
\end{threeparttable}
\end{table*}

Table.\ref{tabsick} shows the average precision for identifying sick chickens with various types of sickness, including frailty, fearfulness, and sex stunting syndrome. Out of a total of 1942 sick chicken targets, there are 697 instances of frailty, 814 instances of fearfulness, and 431 instances of sex stunting syndrome. For each model, the average precision for each type of sickness exceeds 94 $\%$. Although the proposed loss function slightly reduces the average precision for sick categories due to the increased focus on the healthy category during training, which leads to a significant improvement in average precision for healthy chickens (see Table.\ref{tab1}), the proposed knowledge distillation scheme effectively enhances the average precision across all types of sickness.

\subsection{Comparison with existing detectors}
\label{sec3.3}
In this section, we compare the performance of our proposed FCOS-Lite detector with that of several classic and light-weighted detectors, including two of the smallest YOLOv5 \cite{ex1} models (YOLOv5n and YOLOv5s), SSD-Lite (Single Shot Multibox Detector \cite{ex2}) with a MobileNetV2 backbone, and two of the smallest models of another anchor-free detector YOLOX \cite{ex3} (-Nano and -Tiny). It is important to note that, for a fair comparison, the hyper-parameters of the compared detectors are meticulously tuned to ensure optimal performance. Additionally, the input data size for all detectors is standardized to $320 (H)\times320 (W)\times 3 (C)$, with the exception of SSD-Lite, which has an input data size of $300 (H)\times300 (W)\times 3 (C)$. The accuracy metrics, such as mAP@50, AP-sick, Precision (P), and F1-scores, for all detectors are evaluated using PyTorch. However, to determine the effectiveness of deploying the detector on our edge device (IMX500 CMOS sensor), the model sizes of all detectors are compared using TF-Lite after converting the model format from PyTorch to TFLite and performing int8 quantization.  

\begin{table*}[h]
\caption{Comparison of different detectors.}\label{tab3}
\begin{threeparttable}
\begin{tabular*}{\textwidth}{@{\extracolsep\fill}lcccccccc}
\toprule%
Model & Size\tnote{a}& Params& FLOPs& mAP@0.5& AP-sick& P& F1 \\ %
 &(M) & (M)  & (B) & ($\%$) & ($\%$) & ($\%$) & ($\%$) \\
\midrule
 SSD-Lite\tnote{b}& 4.3 & 3.41 & 4.11 & 75.1 & 82.4 & 74.4 & 74.1  \\\midrule
 YOLOX-Nano& 1.2 & 0.91 & 1.08  & 82.3 & 89.7 & 81.9 & 82.1 \\ 
 YOLOX-Tiny& 5.2\tnote{d} & 5.06 & 6.45  & 88.9 & 95.4 & 90.3 & 88.5   \\ \midrule
 YOLOv5n\tnote{c}& 2.1 & 1.68 & 4.5 & 85.8  & 92.7 & 85.1 & 84.4 \\
 YOLOv5s& 8.0\tnote{d} & 6.69 & 16.5 &91.7  & 96.4 & 90.9 & 90.1 \\\midrule
 Ours & 3.3 & 2.84 & 1.95 & \textbf{95.1} & 98.1 & 94.3 & \textbf{94.2}\\
\bottomrule
\end{tabular*}
\begin{tablenotes}
\item[a]{Sizes of all models are evaluated using TFLite after int8 quantization.}
\item[b]{Backbone is MobileNetV2, width $\times$ height of input data size is 300 $\times$ 300.}
\item[c]{Version of YOLOv5 is 6}
\item[d]{Fitting into the edge device is deemed difficult if the TFLite model size exceeds 5 MB.}
\end{tablenotes}
\end{threeparttable}
\end{table*}

\begin{figure*}[htbp]
\centering
\captionsetup[subfigure]{labelformat=empty}
\makebox[\textwidth][c]{
\begin{subfigure}
\centering
{\includegraphics[width=\textwidth]{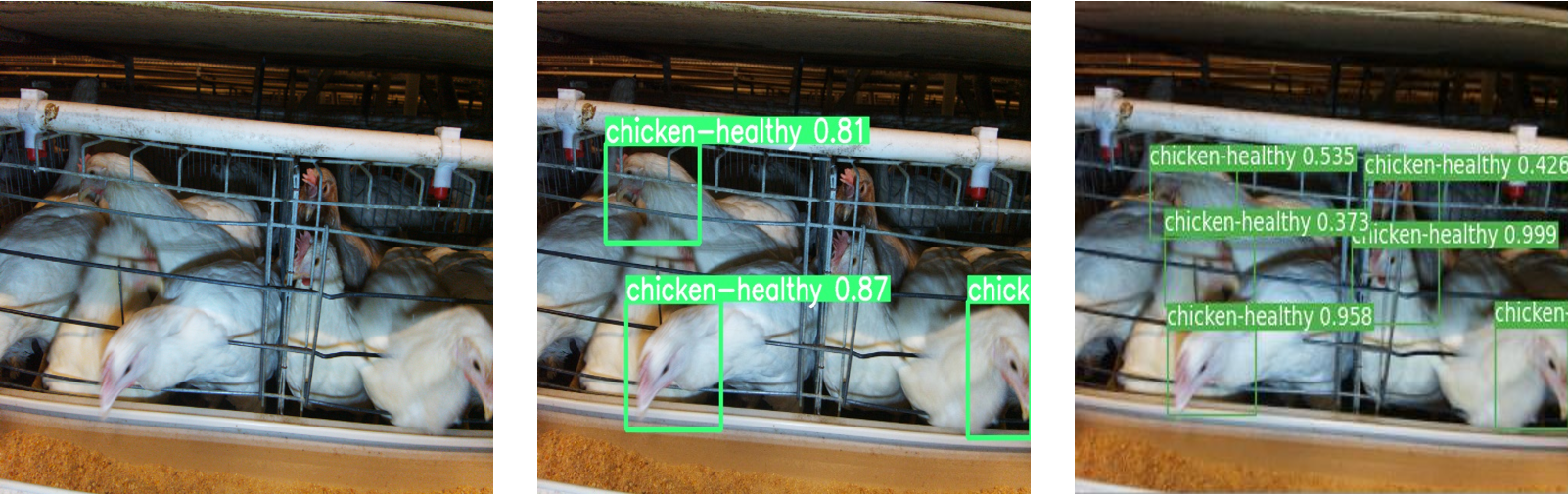}}
\end{subfigure}
}
\makebox[\textwidth][c]{
\begin{subfigure}
\centering
{\includegraphics[width=\textwidth]{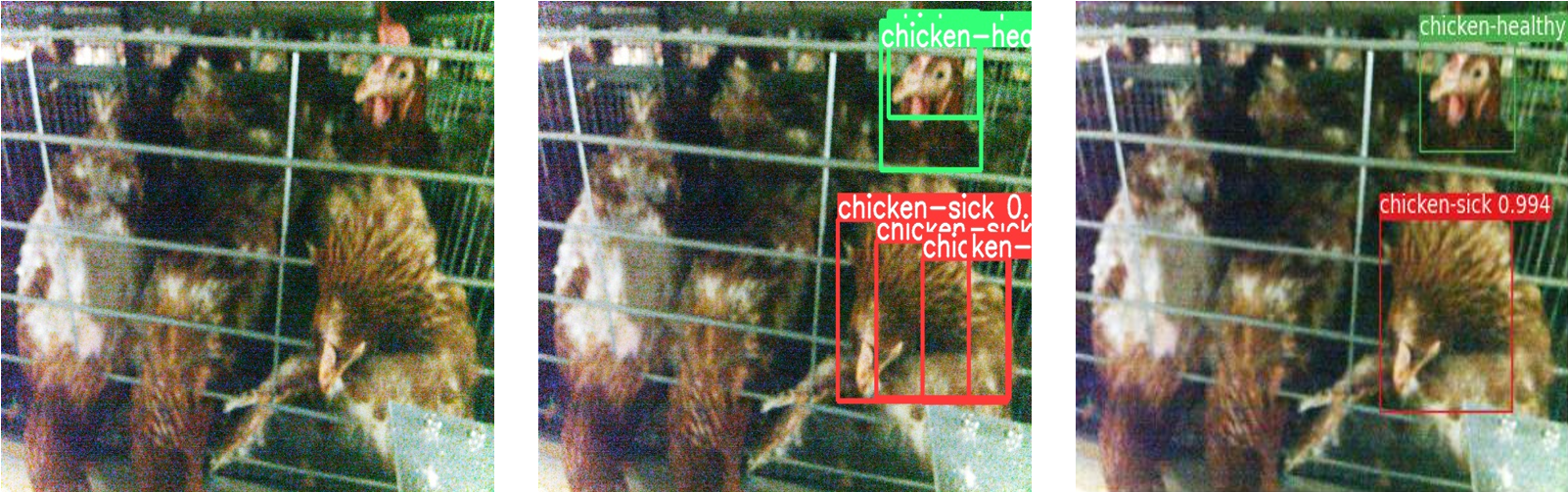}}
\end{subfigure}
}
\makebox[\textwidth][c]{
\begin{subfigure}
\centering
{\includegraphics[width=\textwidth]{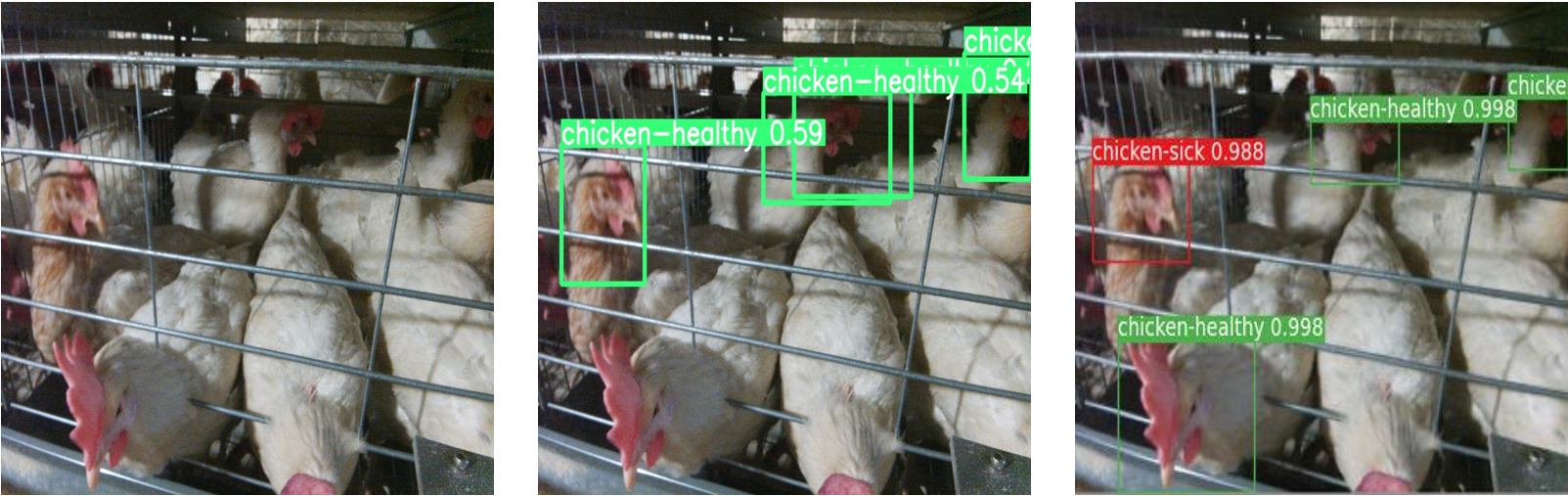}}
\end{subfigure}
}
\makebox[\textwidth][c]{
\begin{subfigure}
\centering
{\includegraphics[width=\textwidth]{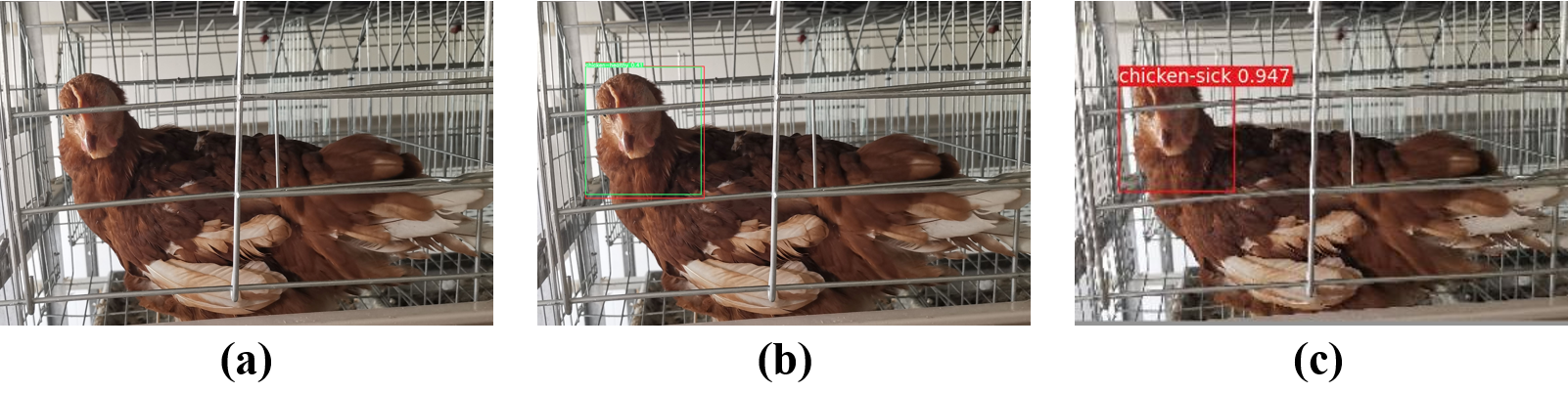}}
\end{subfigure}
}

\caption{Visual comparison between the results obtained from YOLOv5s (b) and our detector (c), with the corresponding input images (a) for clear verification.}
\label{fig8}
\end{figure*}

Table.\ref{tab3} shows the comparative results between our detector and other light-weighted detectors. Due to space constraints, we focus on displaying the metrics for mAP@0.5, AP of the "sick" category, precision (P), and F1-score. As shown in Table.\ref{tab3}, our proposed FCOS-Lite detector outperforms other light-weighted detectors. Compared to models with smaller sizes, such as YOLOX-Nano and YOLOv5n, our detector has achieves approximately 12 $\%$ and 10 $\%$ higher accuracy, respectively. On the other hand, compared to models that exceed the size of our detector, such as SSD-Lite, YOLOX-Tiny, and YOLOv5s, our detector achieves approximately 19 $\%$, 4.7 $\%$ and 3.2 $\%$ higher accuracy, respectively. Notably, compared to YOLOv5s, which has the highest accuracy among existing detectors, our proposed FCOS-Lite detector maintains a model size that is less than half as large while still achieving improvements of 3.4$\%$, 1.7$\%$ and 4.1$\%$ in mAP@0.5, AP for sick chicken category and F1-score, respectively. It is important to note that, although the edge device (IMX500) has a total memory size of 8 MB, the allocated memory space for the AI model and inference should be kept under 5 MB to ensures that other tasks such as image capturing and processing, post-processing, etc., can run smoothly. Consequently, deploying YOLOv5s and YOLOX-Tiny onto the CMOS sensor is challenging due to their larger model sizes. In contrast, our proposed detector not only outperforms these models but is also easily deployable on the CMOS sensor, meeting the memory constraints effectively. We also compared the inference time of all detectors, finding that our proposed detector achieves a real-time performance level of 24 ms (41 FPS) on the PyTorch platform.

Fig.\ref{fig8} shows the visual comparison between the results from YOLOv5s (b) and our detector (c). Due to space constraints, only the visual results of YOLOv5s are presented as the competitor, as it demonstrates the best performance among the classic detectors. The corresponding input images are shown in (a) for easy verification. In (b) and (c), the green boxes and red boxes denote the detected bounding boxes of healthy chickens and sick chickens, respectively. In comparison to YOLOv5s, our detector demonstrates a lower missed detection rate (refer to the top row) and higher accuracy in both localization (refer to the 2$^{nd}$ and 3$^{rd}$ rows) and classification (refer to the bottom two rows). It's noteworthy that in the image of the 2$^{nd}$ row, the sick status of the chicken is indicated by fearfulness. Similarly, in the images of the bottom two rows, the sick statuses of the chickens are indicated by sex stunting syndrome.

\begin{table*}[h]
\caption{Accuracy comparison between the model before quantization (PyTorch Ver.) and the model (Edge-AI Ver.) deployed onto edge-AI CMOS sensor after quantization.}\label{tab4}
\resizebox{\textwidth}{!}{
\begin{threeparttable}
\begin{tabular*}{1.15\textwidth}{@{\extracolsep\fill}lcccccccccc}
\toprule%
Ver. &  Size / Memory  &FPS & mAP@0.5& AP-sick& AP-healthy& P& F1 \\ %
  & requirement (MB)  &  & ($\%$) & ($\%$) & ($\%$) & ($\%$)  &($\%$)\\
\midrule
 PyTorch\tnote{a}& \textcolor{red}{11.7} / 5 &-- & 95.1 & 98.1 & 92.1 & 94.3  & 94.2  \\
 Edge-AI& \textbf{3.3} / 5 &\textbf{27}& \textbf{94.3} & 97.4 & 91.2   & 93.4 & \textbf{93.3}  \\ \midrule
 Diff. $\downarrow$\tnote{b}& 8.4 & -- & \textbf{0.8} & 0.7& 0.9 & 1.1  & \textbf{0.9}   \\
\bottomrule
\end{tabular*}
\begin{tablenotes}
\item[a]{PyTorch is only used for Training AI model and no inference time measurement needed.}
\item[b]{Accuracy decrease value from PyTorch model to Edge-AI model due to int8 quantization, $\downarrow$ indicates that a lower value is better.}
\end{tablenotes}
\end{threeparttable}
}
\end{table*}
\subsection{Implementation on edge device}
\label{sec3.4}
After int8 quantization using TFLite, our FCOS-Lite detector's model size was reduced from 11.7 MB to 3.3 MB, making it compatible with the memory requirements of the edge-AI enabled CMOS sensor - IMX500 (8 MB total memory space, in which at most 5 MB allocated for model inference). We deployed the model onto the CMOS sensor, and conducted a comprehensive evaluation using our test dataset. This evaluation was conducted to quantify the decrease in accuracy from the trained AI model (PyTorch version) to the on-edge deployment model achieved through model quantization. It also aimed to measure the inference time on the edge-AI enabled CMOS sensor. For easy performance measurement of the CMOS sensor, we utilized a Raspberry Pi 4B board connected to the CMOS sensor via CSI connection, as our testing machine. Following this setup, we ran the testing program on the Raspberry Pi to acquire reports on the proposed detector's operational accuracy and inference time from the edge-AI CMOS sensor. Table.\ref{tab4} shows the comparative accuracy results between the PyTorch version model (Float32) and the edge-AI model (int8). Notably, the final accuracy achieved on the edge CMOS sensor still remains commendable, surpassing that of other light-weighted detectors. Only a marginal decrease in accuracy is observed for the edge-AI model. This decrease is caused by the integration of float format data during model quantization and falls within a slight range, typically ranging from approximately 0.8 $\sim$ 1.1$\%$ for each metric. Additionally, the average inference time of the edge-AI model deployed on the CMOS sensor is approximately 37 ms, translating to about 27 frames per second (FPS) in real scenario applications. In conclusion, our proposed FCOS-Lite detector, deployed on the edge-AI enabled CMOS sensor IMX500, demonstrates high detection accuracy in real-time processing, thus proving its suitability for AIoT scenarios.

\begin{figure*}[htbp]%

\centering
\captionsetup[subfigure]{labelformat=empty}
\makebox[\textwidth][c]{
\begin{subfigure}
\centering
{\includegraphics[width=\textwidth]{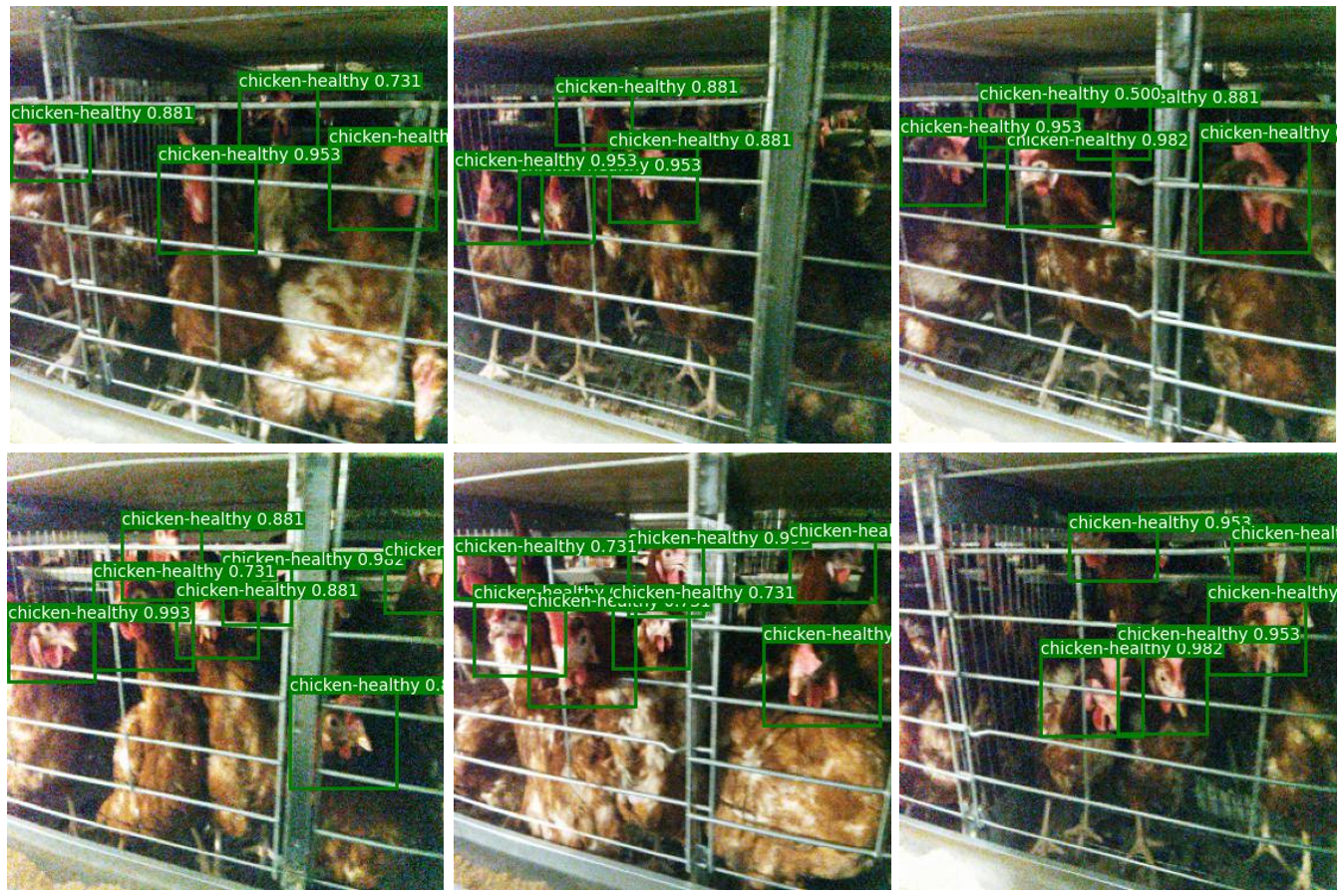}}
\end{subfigure}
}
\makebox[\textwidth][c]{
\begin{subfigure}
\centering
{\includegraphics[width=\textwidth]{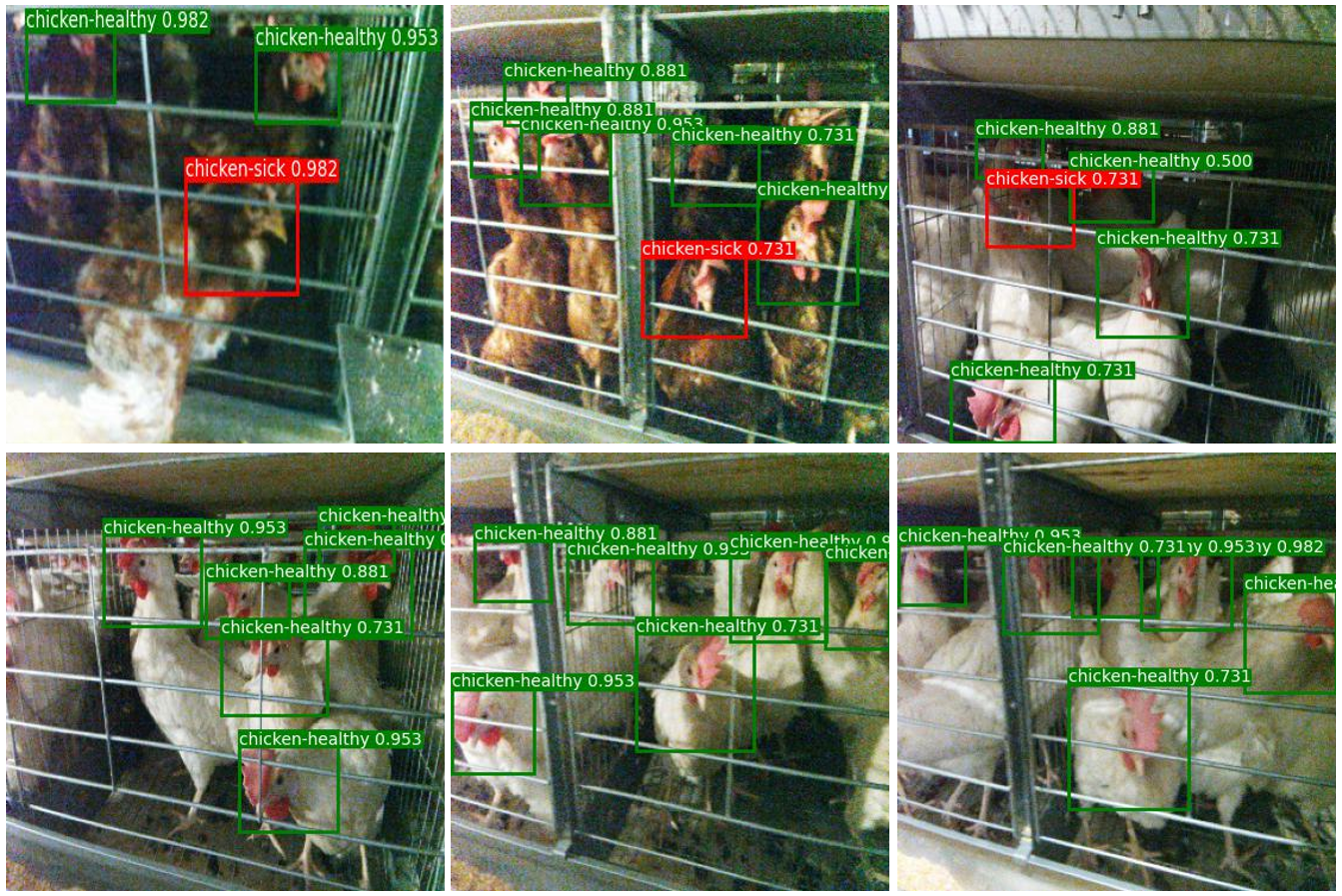}}
\end{subfigure}
}
\caption{Visual results pf the proposed detector deployed on the edge-AI enabled CMOS sensor during the system with intelligent cameras patrolling in real AIoT scenario.}\label{fig11}
\end{figure*}

Fig.\ref{fig11} shows some visual results of the proposed detector operating on the edge-AI enabled CMOS sensor within the system with intelligent cameras patrolling in real AIoT scenario. The visual result showcases the real-time functionality and high accuracy achieved by our proposed edge-AI enabled detector when deployed on the CMOS sensor. Chickens exhibiting "unhealthy" conditions were accurately identified (as indicated by the red boxes), attributed to their abnormal appearances. Meanwhile, the remaining chickens were correctly detected as healthy (as indicated by the green boxes). Additionally, our proposed detector demonstrates a certain of robustness in handling input images with noise and poor lighting conditions in real scenarios. However, to ensure the applicability of our proposed detector across various chicken houses, enhancing its robustness to variable lighting conditions and low image quality will be a focus of our future work. Furthermore, the ambiguity between "healthy" and "unhealthy" chickens based on their appearances poses a challenge (refer to the right image at the 3$^{rd}$ row), potentially affecting the final accuracy of the proposed detector. Therefore, effectively distinguishing between genuine "healthy" and "unhealthy" chickens will remain another focus of our future work. 

\subsection{Limitations and discussion}
\label{sec3.5}

From the evaluation results, we observed a disparity (more than 6$\%$) in accuracy between "AP-sick" and "AP-healthy", representing the classification accuracy for sick and healthy chickens, respectively. This disparity is caused by an imbalanced distribution of attributes, where the appearances and postures of healthy chickens are more varied compared to those of sick chickens, thereby reducing the classification and localization accuracy. Hence, enhancing the discriminatory power and robustness of the detector to differentiate between "healthy" and "unhealthy" chickens based on their visual appearances, as well as handling low-quality input images, are other areas for our further research.

Additionally, sample imbalance is another common challenge  in chicken health status recognition. In real-world scenarios, healthy chickens typically far outnumber sick chickens, which can sometimes lead to biased model training towards the healthy chicken class, ultimately resulting in poor accuracy and generalization, particularly for the sick chicken class. Techniques such as oversampling, weighted sampling, and data augmentation may be applied in our future work to address this challenge. 

Another challenge is improving the generality of the proposed method for varying environmental conditions. For instance, different layer houses may have varied lighting conditions and diverse chicken types, which may limit the performance of our proposed model. Hence, in future work, we will utilize data augmentation methods, regularization techniques, transfer learning, multi-task learning, ensemble methods, and other strategies to enhance the generality and robustness of our proposed model. 

So far, we have only applied Post-Training Quantization (PTQ) for model quantization due to its simplicity and minimal accuracy drops for FCOS and FCOS-Lite models. However, Quantization-Aware Training (QAT) generally maintains higher accuracy and robustness during quantization. Additionally, some advanced techniques such as non-uniform weight quantization \cite{ex11} and sub-byte quantization \cite{ex12}, etc. are also beneficial for enhancing accuracy and reducing memory footprint. Therefore, exploring more effective quantization methods for our models is a future research direction.

In this work, we selected knowledge distillation as technical route. However, from a model size perspective, model pruning especially aggressive pruning, can achieve a higher model compression rate. Therefore, exploring and integrating some advanced pruning techniques such as channel pruning \cite{ex14}, automatic sparse connectivity learning \cite{ex13}, etc. with our proposed method will be our future research endeavors. This approach holds promise for developing a significantly smaller model-sized detector with excellent accuracy for edge-AI enabled devices.

In addition to the aforementioned future endeavors, our next focus is on integrating multiple sources of information, including images of chicken bodies, environmental indices such as temperature and humidity, and auditory cues from chicken sounds. This holistic approach has the potential to furnish an automated system based on AIoT devices with comprehensive information for accurate identification of chicken health statuses.
\vspace{-4pt}
\section{Conclusion}
\label{sec4}
In this study, we have introduced a real-time and compact edge-AI enabled detector tailored for the recognition of healthy statuses in chickens. Unlike other detectors, our solution is specifically designed for deployment on intelligent cameras equipped with edge-AI enabled CMOS sensors. To address the challenges posed by memory constraints in these sensors, we introduced a edge-AI enabled FCOS-Lite detector that utilizes MobileNet as the backbone and integrates modified neck and head components, ensuring a compact model size. Furthermore, to mitigate the accuracy reduction typically associated with compact edge-AI models, we proposed novel approaches such as a gradient weighting loss function and CIoU loss function as classification loss and localization loss, respectively. Additionally, a knowledge distillation scheme was employed to transfer valuable insights from a larger teacher detector to our FCOS-Lite detector, thereby enhancing performance while maintaining a compact model size and the same inference costs. Our experimental results validate the effectiveness of our proposed approach, with the edge-AI enabled model achieving commendable performance metrics. Specifically, using the trained AI model with PyTorch, we attained an mAP of 95.1$\%$ and an F1-score of 94.2$\%$, and deploying the model on the edge-AI enabled CMOS sensor yielded an mAP of 94.3$\%$ and an F1-score of 93.3$\%$. Moreover, the detector operates efficiently at a speed exceeding 20 FPS on the edge-AI enabled CMOS sensor. These findings highlight the practical feasibility of our solution for automated poultry health monitoring. By utilizing lightweight intelligent cameras with minimal power consumption and bandwidth costs, our approach ensures practical applicability in AIoT-based smart poultry farming scenarios.
\vspace{-4pt}
\subsection*{Acknowledgements} 
This research was supported by SONY Research Foundation and conducted at SONY Research $\&$ Development Center China, Beijing Lab.

\bibliographystyle{elsarticle-num} 
\bibliography{mybibfile}

\begin{thebibliography}{10}
\expandafter\ifx\csname url\endcsname\relax
  \def\url#1{\texttt{#1}}\fi
\expandafter\ifx\csname urlprefix\endcsname\relax\def\urlprefix{URL }\fi
\expandafter\ifx\csname href\endcsname\relax
  \def\href#1#2{#2} \def\path#1{#1}\fi

\bibitem{intro1}
R.~O. Ojo, A.~O. Ajayi, H.~A. Owolabi, L.~O. Oyedele, L.~A. Akanbi, Internet of things and machine learning techniques in poultry health and welfare management: A systematic literature review, Computers and Electronics in Agriculture 200 (2022) 107266, \url{https://doi.org/10.1016/j.compag.2022.107266}.

\bibitem{intro3}
A.~Banakar, M.~Sadeghi, A.~Shushtari, An intelligent device for diagnosing avian diseases: Newcastle, infectious bronchitis, avian influenza, Computers and electronics in agriculture 127 (2016) 744--753, \url{https://doi.org/10.1016/j.compag.2016.08.006}.

\bibitem{intro2}
O.~Debauche, S.~Mahmoudi, S.~A. Mahmoudi, P.~Manneback, J.~Bindelle, F.~Lebeau, Edge computing and artificial intelligence for real-time poultry monitoring, Procedia computer science 175 (2020) 534--541, \url{https://doi.org/10.1016/j.procs.2020.07.076}.

\bibitem{intro4}
G.~Ren, T.~Lin, Y.~Ying, G.~Chowdhary, K.~Ting, Agricultural robotics research applicable to poultry production: A review, Computers and Electronics in Agriculture 169 (2020) 105216, \url{https://doi.org/10.1016/j.compag.2020.105216}.

\bibitem{intro19}
W.~F. Pereira, L.~da~Silva~Fonseca, F.~F. Putti, B.~C. Góes, L.~de~Paula~Naves, Environmental monitoring in a poultry farm using an instrument developed with the internet of things concept, Computers and Electronics in Agriculture 170 (2020) 105257, \url{https://doi.org/10.1016/j.compag.2020.105257}.

\bibitem{intro5}
A.~A.~G. Raj, J.~G. Jayanthi, Iot-based real-time poultry monitoring and health status identification, in: 2018 11th International Symposium on Mechatronics and its Applications (ISMA), 2018, pp. 1--7, \url{https://doi.org/10.1109/ISMA.2018.8330139}.

\bibitem{intro6}
A.~G.~R. Alex, G.~J. Joseph, Real-time poultry health identification using iot test setup, optimization and results, in: Advances in Signal Processing and Intelligent Recognition Systems: 4th International Symposium SIRS 2018, Bangalore, India, September 19--22, 2018, Revised Selected Papers 4, 2019, pp. 30--40, \url{https://doi.org/10.1007/978-981-13-5758-9_3}.

\bibitem{iot2}
SONY, {Overview - IMX500}, \url{https://developer.sony.com/imx500/} (2023).

\bibitem{intro17}
S.~Cakic, T.~Popovic, S.~Krco, D.~Nedic, D.~Babic, I.~Jovovic, Developing edge ai computer vision for smart poultry farms using deep learning and hpc, Sensors 23~(6) (2023) 3002, \url{ https://doi.org/10.3390/s23063002}.

\bibitem{intro10}
Z.~S. Yi~Shi, Shen~Lian, et~al., Recognition method of pheasant using enhanced tiny-yolov3 model, Transactions of the Chinese Society of Agricultural Engineering 13 (2020) 141--147, \underline{10.11975/j.issn.1002-6819.2020.13.017}.

\bibitem{intro11}
X.~Zhuang, T.~Zhang, Detection of sick broilers by digital image processing and deep learning, Biosystems Engineering 179 (2019) 106--116, \url{https://doi.org/10.1016/j.biosystemseng.2019.01.003}.

\bibitem{intro12}
H.-W. Liu, C.-H. Chen, Y.-C. Tsai, K.-W. Hsieh, H.-T. Lin, Identifying images of dead chickens with a chicken removal system integrated with a deep learning algorithm, Sensors 21~(11) (2021) 3579, \url{https://doi.org/10.3390/s21113579}.

\bibitem{intro13}
H.~Hao, P.~Fang, E.~Duan, Z.~Yang, L.~Wang, H.~Wang, A dead broiler inspection system for large-scale breeding farms based on deep learning, Agriculture 12~(8) (2022) 1176, \url{https://doi.org/10.3390/agriculture12081176}.

\bibitem{intro14}
Q.~Tong, E.~Zhang, S.~Wu, K.~Xu, C.~Sun, A real-time detector of chicken healthy status based on modified yolo, Signal, Image and Video Processing 17~(8) (2023) 4199--4207, \url{https://doi.org/10.1007/s11760-023-02652-6}.

\bibitem{intro20}
J.~Yang, T.~Zhang, C.~Fang, H.~Zheng, A defencing algorithm based on deep learning improves the detection accuracy of caged chickens, Computers and Electronics in Agriculture 204 (2023) 107501, \url{https://doi.org/10.1016/j.compag.2022.107501}.

\bibitem{intro21}
M.~Campbell, P.~Miller, K.~Díaz-Chito, X.~Hong, N.~McLaughlin, F.~Parvinzamir, J.~{Martínez Del Rincón}, N.~O'Connell, A computer vision approach to monitor activity in commercial broiler chickens using trajectory-based clustering analysis, Computers and Electronics in Agriculture 217 (2024) 108591, \url{https://doi.org/10.1016/j.compag.2023.108591}.

\bibitem{intro22}
A.~Nasiri, Y.~Zhao, H.~Gan, Automated detection and counting of broiler behaviors using a video recognition system, Computers and Electronics in Agriculture 221 (2024) 108930, \url{https://doi.org/10.1016/j.compag.2024.108930}.

\bibitem{intro23}
X.~Tan, C.~Yin, X.~Li, M.~Cai, W.~Chen, Z.~Liu, J.~Wang, Y.~Han, Sy-track: A tracking tool for measuring chicken flock activity level, Computers and Electronics in Agriculture 217 (2024) 108603, \url{https://doi.org/10.1016/j.compag.2023.108603}.

\bibitem{intro16}
Y.~Guo, Z.~Yu, Z.~Hou, W.~Zhang, G.~Qi, Sheep face image dataset and dt-yolov5s for sheep breed recognition, Computers and Electronics in Agriculture 211 (2023) 108027, \url{https://doi.org/10.1016/j.compag.2023.108027}.

\bibitem{intro24}
N.~Rai, Y.~Zhang, M.~Villamil, K.~Howatt, M.~Ostlie, X.~Sun, Agricultural weed identification in images and videos by integrating optimized deep learning architecture on an edge computing technology, Computers and Electronics in Agriculture 216 (2024) 108442, \url{https://doi.org/10.1016/j.compag.2023.108442}.

\bibitem{intro18}
D.~Mao, H.~Sun, X.~Li, X.~Yu, J.~Wu, Q.~Zhang, Real-time fruit detection using deep neural networks on cpu (rtfd): An edge ai application, Computers and Electronics in Agriculture 204 (2023) 107517, \url{https://doi.org/10.1016/j.compag.2022.107517}.

\bibitem{dl1}
Z.~Tian, C.~Shen, H.~Chen, T.~He, Fcos: Fully convolutional one-stage object detection, in: Proceedings of the IEEE/CVF international conference on computer vision, 2019, pp. 9627--9636, \url{ https://doi.org/10.48550/arXiv.1904.01355}.

\bibitem{intro15}
Z.~Jiao, K.~Huang, G.~Jia, H.~Lei, Y.~Cai, Z.~Zhong, An effective litchi detection method based on edge devices in a complex scene, Biosystems Engineering 222 (2022) 15--28, \url{https://doi.org/10.1016/j.biosystemseng.2022.07.009}.

\bibitem{dl3}
K.~He, X.~Zhang, S.~Ren, J.~Sun, Deep residual learning for image recognition, in: Proceedings of the IEEE conference on computer vision and pattern recognition, 2016, pp. 770--778, \url{https://doi.org/10.1109/CVPR.2016.90}.

\bibitem{dl2}
A.~G. Howard, M.~Zhu, B.~Chen, D.~Kalenichenko, W.~Wang, T.~Weyand, M.~Andreetto, H.~Adam, Mobilenets: Efficient convolutional neural networks for mobile vision applications, arXiv preprint arXiv:1704.04861\url{ https://doi.org/10.48550/arXiv.1704.04861} (2017).

\bibitem{dl6}
T.-Y. Lin, P.~Goyal, R.~Girshick, K.~He, P.~Doll{\'a}r, Focal loss for dense object detection, in: Proceedings of the IEEE international conference on computer vision, 2017, pp. 2980--2988, \url{ https://doi.org/10.48550/arXiv.1708.02002}.

\bibitem{dl7}
Z.~Zheng, P.~Wang, W.~Liu, J.~Li, R.~Ye, D.~Ren, Distance-iou loss: Faster and better learning for bounding box regression, in: Proceedings of the AAAI conference on artificial intelligence, Vol.~34, 2020, pp. 12993--13000, \url{https://doi.org/10.1609/aaai.v34i07.6999}.

\bibitem{dl8}
J.~Yu, Y.~Jiang, Z.~Wang, Z.~Cao, T.~Huang, Unitbox: An advanced object detection network, in: Proceedings of the 24th ACM international conference on Multimedia, 2016, pp. 516--520, \url{https://doi.org/10.1145/2964284.2967274}.

\bibitem{lo1}
H.~Rezatofighi, N.~Tsoi, J.~Gwak, A.~Sadeghian, I.~Reid, S.~Savarese, Generalized intersection over union: A metric and a loss for bounding box regression, in: Proceedings of the IEEE/CVF conference on computer vision and pattern recognition, 2019, pp. 658--666, \url{https://doi.org/10.48550/arXiv.1902.09630}.

\bibitem{kd17}
Z.~Yang, Z.~Li, X.~Jiang, Y.~Gong, Z.~Yuan, D.~Zhao, C.~Yuan, Focal and global knowledge distillation for detectors, in: Proceedings of the IEEE/CVF Conference on Computer Vision and Pattern Recognition, 2022, pp. 4643--4652, \url{ https://doi.org/10.48550/arXiv.2111.11837}.

\bibitem{ex1}
G.~Jocher, {YOLOv5}, \url{https://github.com/ultralytics/yolov5} (2022).

\bibitem{ex2}
W.~Liu, D.~Anguelov, D.~Erhan, C.~Szegedy, S.~Reed, C.-Y. Fu, A.~C. Berg, Ssd: Single shot multibox detector, in: ECCV, 2016, pp. 21--37, \url{https://doi.org/10.1007/978-3-319-46448-0_2}.

\bibitem{ex3}
Z.~Ge, S.~Liu, F.~Wang, Z.~Li, J.~Sun, Yolox: Exceeding yolo series in 2021, arXiv preprint arXiv:2107.08430\url{ https://doi.org/10.48550/arXiv.2107.08430} (2021).

\bibitem{ex11}
Q.~Huang, Z.~Tang, High-performance and lightweight ai model for robot vacuum cleaners with low bitwidth strong non-uniform quantization, AI 4~(3) (2023) 531--550, \url{https://doi.org/https://doi.org/10.3390/ai4030029}.

\bibitem{ex12}
E.~Park, S.~Yoo, Profit: A novel training method for sub-4-bit mobilenet models, in: Computer Vision--ECCV 2020: 16th European Conference, Glasgow, UK, August 23--28, 2020, Proceedings, Part VI 16, Springer, 2020, pp. 430--446, \url{https://doi.org/10.48550/arXiv.2008.04693}.

\bibitem{ex14}
W.~Hu, Z.~Che, N.~Liu, M.~Li, J.~Tang, C.~Zhang, J.~Wang, Catro: Channel pruning via class-aware trace ratio optimization, IEEE Transactions on Neural Networks and Learning Systems\url{https://doi.org/10.48550/arXiv.2110.10921} (2023).

\bibitem{ex13}
Z.~Tang, L.~Luo, B.~Xie, Y.~Zhu, R.~Zhao, L.~Bi, C.~Lu, Automatic sparse connectivity learning for neural networks, IEEE Transactions on Neural Networks and Learning Systems 34~(10) (2022) 7350--7364, \url{https://doi.org/10.48550/arXiv.2201.05020}.

\end{thebibliography}







\end{document}